\documentclass[preprint, review,3p,times, 10pt]{elsarticle}

\usepackage{caption}
\usepackage{subcaption}

\usepackage{cite}
\usepackage{amsmath,amssymb,amsfonts}
\usepackage{algorithmic}
\usepackage{graphicx}
\usepackage{textcomp}
\usepackage{comment}
\usepackage{footnote}
\usepackage{comment}
\usepackage{threeparttable}
\usepackage{romannum}
\usepackage{longtable}
\usepackage{lscape}
\usepackage{soul}
\usepackage{array, makecell}
\usepackage{soul}
\usepackage{color}
\usepackage[dvipsnames]{xcolor}
\usepackage{longtable}
\usepackage{subcaption}
\usepackage{float}
\usepackage{ threeparttablex}
\usepackage{booktabs}
\usepackage[hyphens]{url}
\usepackage{hyperref}
\hypersetup{colorlinks=true,breaklinks=true}

\journal{Elsevier}
\usepackage[utf8]{inputenc} 
\usepackage[T1]{fontenc}    

\usepackage{multirow}
\usepackage{multicol}
\begin{document}

\begin{frontmatter}
\title{BeliN: A Novel Corpus for Bengali Religious News Headline Generation using Contextual Feature Fusion}

\author[1,2]{Md Osama}\ead{osama_cse@baust.edu.bd}
\author[1]{Ashim Dey}\ead{ashim@cuet.ac.bd}
\author[1]{Kawsar Ahmed}\ead{u1804017@student.cuet.ac.bd}
\author[3]{Muhammad Ashad Kabir\corref{correspondingauthor}}\ead{akabir@csu.edu.au}

\cortext[correspondingauthor]{Corresponding author: Charles Sturt University, Panorama Ave, Bathurst, NSW 2795. Ph.+61263386259}

\affiliation[1]{organization={Department of Computer Science and Engineering, Chittagong University of Engineering and Technology}, city={Raozan}, state={Chittagong}, postcode={4349}, country={Bangladesh}}

\affiliation[2]{organization={Department of Computer Science and Engineering, Bangladesh Army University of Science and Technology (BAUST)}, city={Saidpur}, state={Nilphamari}, postcode={5310}, country={Bangladesh}}

\affiliation[3]{organization={School of Computing, Mathematics and Engineering, Charles Sturt University}, city={Bathurst}, state={NSW}, postcode={2795}, country={Australia}}

\begin{abstract}
Automatic text summarization, particularly headline generation, remains a critical yet underexplored area for Bengali religious news. Existing approaches to headline generation typically rely solely on the article content, overlooking crucial contextual features such as sentiment, category and aspect. This limitation significantly hinders their effectiveness and overall performance. This study addresses this limitation by introducing a novel corpus, BeliN (Bengali Religious News) -- comprising religious news articles from prominent Bangladeshi online newspapers, and \textit{MultiGen} -- a contextual multi-input feature fusion headline generation approach.
Leveraging transformer-based pre-trained language models such as BanglaT5, mBART, mT5, and mT0, \textit{MultiGen} integrates additional contextual features—including category, aspect, and sentiment—with the news content. This fusion enables the model to capture critical contextual information often overlooked by traditional methods.
Experimental results demonstrate the superiority of \textit{MultiGen} over the baseline approach that uses only news content, achieving a BLEU score of 18.61 and ROUGE-L score of 24.19, compared to baseline approach scores of 16.08 and 23.08, respectively. These findings underscore the importance of incorporating contextual features in headline generation for low-resource languages.
By bridging linguistic and cultural gaps, this research advances natural language processing for Bengali and other underrepresented languages. To promote reproducibility and further exploration, the dataset and implementation code are publicly accessible at \url{https://github.com/akabircs/BeliN}.   
\end{abstract}

\end{frontmatter}


\section{Introduction}
\label{sec:introduction}
A newspaper title serves as an essential element, often shaping the reader's first impression by being both representative of the content and attention-grabbing~\citep{userengagingheadline}. 
Representative headlines play a crucial role in information retrieval systems, which prioritize keywords in headlines to enhance searchability and relevance~\citep{de2012chatter}. Consequently, researchers have extensively explored automated techniques, such as text summarization, to generate compelling and accurate headlines from articles~\citep{koh2022empirical}. 

Text summarization systems can be categorized into two main approaches: extractive and abstractive techniques~\citep{10.1145/3700639}.
Early text summarization techniques predominantly relied on extractive processes, which identify and select significant portions of text, such as key phrases or sentences, directly from the document. These methods generate summaries by reproducing the most critical points verbatim, ensuring fidelity to the original content~\citep{BANERJEE2023103291,app13137620}. 
Abstractive text summarization, a more recent development in the field of Automatic Text Summarization (ATS), generates concise summaries by reformulating key ideas from the original text. Unlike extractive methods, which replicate exact phrases, abstractive summarization produces a condensed script that captures the essence of the document in a clear and coherent manner~\citep{alomari2022deep,10.1145/3700639}. ATS aims to identify and appropriately prioritize the informative components of the source article, making it particularly useful for summarizing blogs, newspapers, and other text-based media~\citep{el2021automatic}. 

Headline generation, a specialized form of text summarization, can also employ both extractive and abstractive approaches. However, compared to the extractive technique, the abstractive method more accurately generates real-world headlines because it captures the underlying meaning and context of the content, allowing for greater flexibility and creativity in rephrasing, while the extractive approach tends to rely on directly selecting portions of the input text~\citep{app14020713}. Traditionally, headlines are generated solely based on the content of the article~\citep{Ayana2017}, a method that, while effective in capturing key information, limits the potential for engaging and compelling headline creation~\citep{hagar2019optimizing}. This approach tends to focus purely on summarizing the main points, often neglecting the broader context or emotional resonance required to capture a reader's attention in today's fast-paced media environment~\citep{banerjee2024language}. Furthermore, while there has been extensive research in headline generation for high-resource languages such as English~\citep{el2021automatic}, relatively few studies have focused on Bengali, particularly in the context of Bengali religious news~\citep{akash-etal-2023-shironaam,SAAD2024110874}. One notable study, Shironaam~\citep{akash-etal-2023-shironaam}, focused on Bengali and incorporated additional contextual information, such as category, in the headline generation process. However, their sample of religious news was very small, and they did not consider other significant contextual elements, such as sentiment and aspect. As of 2024, Bengali is spoken by over 237 million native speakers and an additional 41 million second-language speakers, making it the fifth most spoken native language and the seventh most spoken language worldwide~\citep{statb}. This widespread usage underscores Bengali's significant role in global linguistic diversity.
 
\begin{figure}
\centering
\includegraphics[width=.9\textwidth]{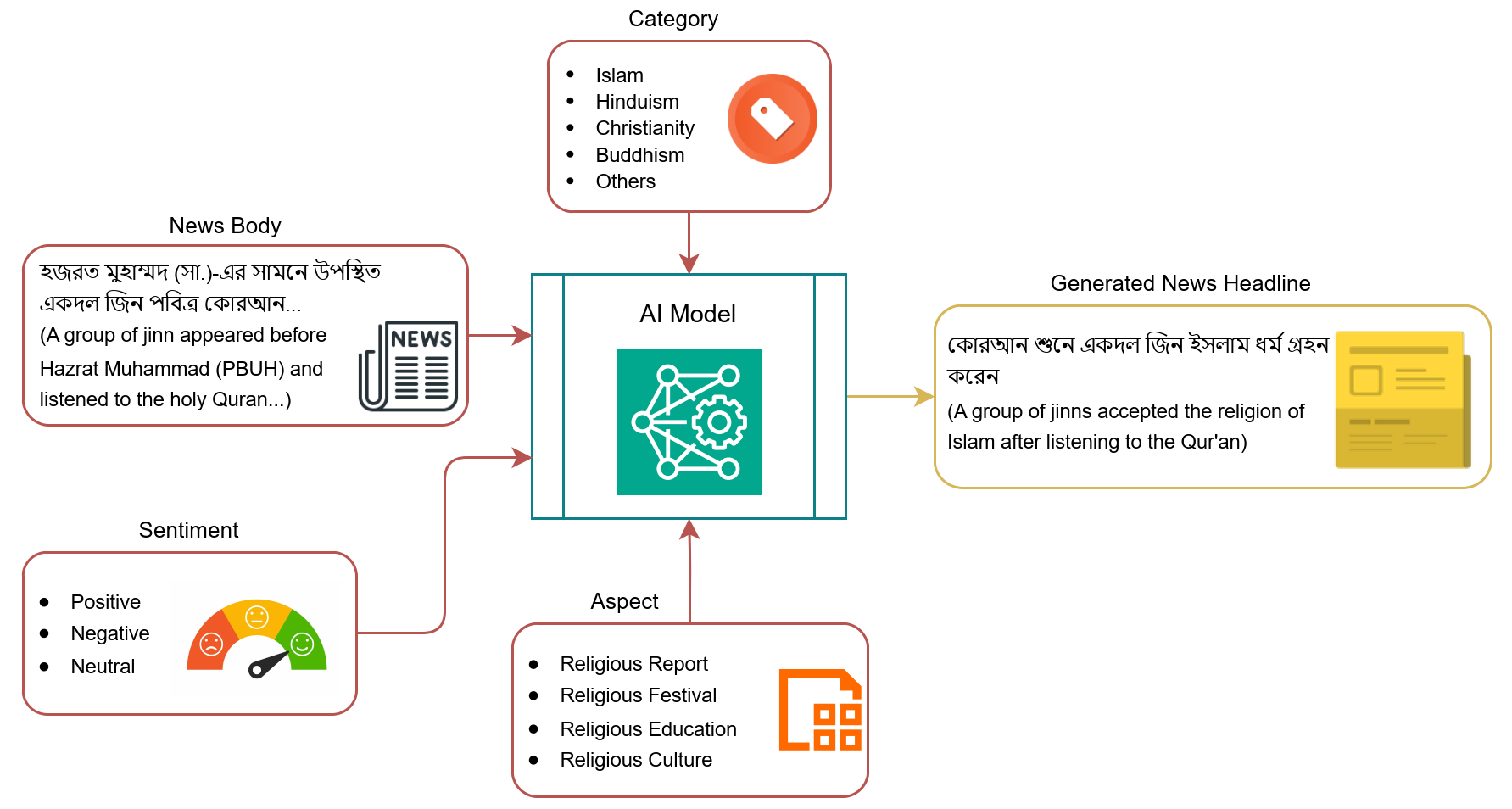}
\caption{An overview of multi-input headline generation
\label{intro}}
\end{figure}

In this paper, we introduce a novel Bengali religious news corpus, named \textit{BeliN}, with a multi-input approach that incorporates additional features alongside the news content to generate more accurate and contextually relevant headlines. Specifically, we include the article’s category, content aspect, and sentiment as additional input features, as illustrated in Figure~\ref{intro}. This approach aims to narrow the domain and enhance the quality of the generated headlines. The rationale behind this method is that a headline should not only capture the essence of the article but also reflect its main context concisely. By integrating multiple inputs, we seek to provide a more comprehensive understanding of the article's content, thereby producing more precise and relevant headlines. The inclusion of these features in the headline generation process is expected to improve the model’s performance and better reflect the underlying article's themes~\citep{akash-etal-2023-shironaam}. This multi-input approach not only narrows the domain but also enriches the generated headlines with additional contextual information, making them more informative and representative of the article.

By leveraging these additional inputs, we aim to create a more robust and accurate headline-generation system for Bengali news articles, particularly within the domain of religious news.
The proposed Bengali news headline generation system, named \textit{MultiGen}, employs and evaluates state-of-the-art pretrained transformer models such as mT5, mT0, mBART, and BanglaT5. Our experimental evaluations show that BanglaT5 outperforms others, offering significant improvements in headline accuracy and contextual relevance.
The key Contributions of this work are as follows:
\begin{itemize}
    \item We have developed a novel dataset, \textit{BeliN}, focused on the underexplored domain of religious news in the low-resource Bengali language. The dataset comprises 2,520 news articles and their corresponding headlines, making it valuable for various NLP tasks, including headline generation, text summarization, news categorization, sentiment analysis and aspect classification.

    \item We have proposed the \textit{MultiGen} approach, which incorporates additional features such as category, aspect, and sentiment as auxiliary information to enhance the headline generation process. Our approach demonstrates the importance of integrating multiple inputs, achieving significant improvements over the baseline approach.

    \item We have employed and evaluated the performance of state-of-the-art pre-trained models for generating compelling and attention-grabbing news headlines.

\end{itemize}
The remainder of this paper is organized as follows: Section \ref{sec:relatedwork} provides a comprehensive review of the related work that forms the foundation of this research. Section \ref{sec:belin} introduces the BeliN corpus, presenting its key characteristics and statistical summary. The methodology and approaches adopted in this study are detailed in Section \ref{sec:multigen}. Section \ref{sec:evaluation} discusses the experimental setup, metrics, and models and presents the evaluation results. A detailed analysis, including a discussion of the findings and limitations, is presented in Section \ref{sec:discussion}. Finally, Section \ref{sec:conclusion} concludes the paper and outlines potential future directions.

\section{Related Work}
\label{sec:relatedwork}
Generating news headlines has been a prominent area of research within the field of natural language processing (NLP)~\citep{servey,zeyad2024advancements}. Although significant advancements have been made in text summarization~\citep{Yadav2023,BharathiMohan2023,cajueiro2023} and headline generation~\citep{Ayana2017,reviewheadline}, progress in low-resource languages, such as Bengali, remains limited~\citep{akash-etal-2023-shironaam,salehin2019generating,hayat2023abstractive,kabir-etal-2024-benllm}. The task of generating headlines, particularly for religious news in Bengali, poses unique challenges due to the scarcity of annotated datasets~\citep{akash-etal-2023-shironaam}.

\begin{table}[!htp]
\centering
\caption{Summary of the related work.}
\begin{threeparttable}
\resizebox{\textwidth}{!}{%
\begin{tabular}{@{\extracolsep{4pt}}cllcccccccc}
\hline
\multirow{2}{*}{Study} & \multicolumn{4}{c}{Dataset/Corpus} & \multicolumn{4}{c}{Feature} & \multirow{2}{*}{Task} & \multirow{2}{*}{Approach} \\  
\cline{2-5} \cline{6-9} 
 & Name & Language & Religious & Availability & Content & Category & Aspect & Sentiment &  & \\ 
\hline\hline
\citep{newsroom} & Newsroom & English & \(\times\) & Public\tnote{a} &\checkmark& \(\times\) & \(\times\) & \(\times\) & S & A \& E \\ 
\citep{sarvanidigital} & CNN Daily Mail & English & \(\times\) & Public\tnote{b} &\checkmark& \(\times\) & \(\times\) & \(\times\) & S & A \\ 
\citep{cnncorpus} & CNN Corpus & English & \(\times\) & Private &\checkmark& \(\times\) & \(\times\) & \(\times\) & S & E \\ 
\citep{jiang-dreyer-2024-ccsum} & CCSum & English & \(\times\) & Public\tnote{c} &\checkmark&\(\times\)&\(\times\)&\(\times\)& S & A \\ 
\citep{xsum-dataset} & XSum & English & \(\times\) & Public\tnote{l} &\checkmark&\(\times\)&\(\times\)&\(\times\)& H & A \\
\citep{representativeheadlines} & NewSHead & English & \(\times\) & Public\tnote{d} &\checkmark&\checkmark&\(\times\)&\(\times\)& H & A \\ 

\citep{pens,putyourvoice} & PENS & English & \(\times\) & Public\tnote{e} &\checkmark&\checkmark&\(\times\)&\(\times\)& H & A \\


\citep{jin2020hooksheadlinelearninggenerate} & CNN, New York Times~\citep{AB2/GZC6PL_2008} & English & \(\times\) & Private &\checkmark&\(\times\)&\(\times\)&\(\times\)& H & A \\ 
\citep{takase2016neural} & DUC~\citep{duc}, Gigaword~\citep{english-gigaword,gigaword} & English & \(\times\) & Public\tnote{f,g} &\checkmark&\(\times\)&\(\times\)&\(\times\)& H & A \\ 

\citep{userengagingheadline} & Newsroom~\citep{newsroom}, Gigaword~\citep{english-gigaword,gigaword} & English & \(\times\) & Public\tnote{a,g} &\checkmark&\(\times\)&\(\times\)&\(\times\)& S & A \& E \\ 
\citep{singh2021sheg} & \makecell[t l]{Newsroom~\citep{newsroom}, Gigaword~\citep{english-gigaword,gigaword},\\ CNN Daily Mail~\citep{sarvanidigital}} & English & \(\times\) & Public\tnote{a,g,b} &\checkmark&\(\times\)&\(\times\)&\(\times\)& S & A \\ 
\citep{suder} & SuDer & Turkish & \(\times\) & Private &\checkmark&\(\times\)&\(\times\)&\(\times\)& H & A \\ 
\citep{ogunremi2024afrihg} & AFRIHG & African & \(\times\) & Private &\checkmark&\(\times\)&\(\times\)&\(\times\)& H & A \\ 
\citep{bukhtiyarov} & RIA~\citep{gavrilov2019self}, Lenta~\citep{lenta} & Russian & \(\times\) & Public\tnote{h,i} &\checkmark&\(\times\)&\(\times\)&\(\times\)& H & A \\ 
\citep{hu-etal-2015-lcsts} & LCSTS & Chinese & \(\times\) & Public\tnote{j} &\checkmark&\(\times\)&\(\times\)&\(\times\)& S & A \\ 

\citep{madasu-etal-2023-mukhyansh} & Mukhyansh & Indic languages & \(\times\) & Private &\checkmark&\(\times\)&\(\times\)&\(\times\)& H & A \\
\citep{aralikatte-etal-2023-varta} & Varta & Indic, English & \(\times\) & Public\tnote{k} &\checkmark&\(\times\)&\(\times\)&\(\times\)& H & A \\ 
\citep{9507422} & LCSTS~\citep{hu-etal-2015-lcsts}, XSum~\citep{xsum-dataset} & Chinese, English & \(\times\) & Public\tnote{j,l} &\checkmark&\(\times\)&\(\times\)&\(\times\)& H & A \\ 
\citep{gavrilov2019self} & RIA, New York Times~\citep{AB2/GZC6PL_2008} & Russian, English & \(\times\) & Private (NYT) &\checkmark&\(\times\)&\(\times\)&\(\times\)& H & A \\ 
\citep{salehin2019generating} & Own dataset & Bengali & \(\times\) & Private &\checkmark&\(\times\)&\(\times\)&\(\times\)& H & A \\ 
\citep{hasan2021xlsumlargescalemultilingualabstractive} & XL-Sum & Bengali+43 others & \(\times\) & Public\tnote{m} &\checkmark&\(\times\)&\(\times\)&\(\times\)& S & A \\ 
\citep{potrika} & Potrika & Bengali & \(\times\) & Public\tnote{n} &\checkmark&\checkmark&\(\times\)&\(\times\)& H & A \\ 
\citep{SAAD2024110874} & BNAD & Bengali & \checkmark & Public\tnote{o} &\checkmark&\checkmark&\(\times\)&\(\times\)& H & A \\ 
\citep{akash-etal-2023-shironaam} & Shironaam & Bengali & \checkmark & Public\tnote{p} &\checkmark&\checkmark&\(\times\)&\(\times\)& H & A \\ \hline
Ours & BeliN & Bengali & \checkmark & Public\tnote{q} &\checkmark&\checkmark&\checkmark&\checkmark& H & A \\ 
\hline

\hline
\multicolumn{11}{l}{S = Summarization, H = Headline, A = Abstractive, E = Extractive}
\end{tabular}%
}
\begin{tablenotes}
\scriptsize
\item[a] \url{https://paperswithcode.com/dataset/newsroom}
\item[b] \url{https://www.kaggle.com/datasets/gowrishankarp/newspaper-text-summarization-cnn-dailymail}
\item[c] \url{https://github.com/amazon-science/ccsum}
\item[d] \url{https://github.com/google-research-datasets/NewSHead}
\item[e] \url{https://msnews.github.io/pens.html}
\item[f] \url{https://www-nlpir.nist.gov/projects/duc/data.html}
\item[g] \url{https://www.kaggle.com/datasets/arngowda/gigaword-corpus}
\item[h] \url{https://github.com/RossiyaSegodnya/ria_news_dataset}
\item[i] \url{https://github.com/yutkin/Lenta.Ru-News-Dataset}
\item[j] \url{https://huggingface.co/datasets/hugcyp/LCSTS}
\item[k] \url{https://github.com/rahular/varta}
\item[l] \url{https://github.com/EdinburghNLP/XSum/tree/master/XSum-Dataset}
\item[m] \url{https://github.com/csebuetnlp/xl-sum}
\item[n] \url{https://doi.org/10.17632/v362rp78dc.4}
\item[o] \url{https://doi.org/10.5281/zenodo.11069882}
\item[p] \url{https://github.com/dialect-ai/BenHeadGen}
\item[q] \url{https://github.com/akabircs/BeliN}
\end{tablenotes}
\end{threeparttable}
\label{tab:tabular_analysis}
\end{table}

In high-resource languages like English and Chinese, a variety of datasets have supported significant advancements in summarization and headline generation research summarized in Table \ref{tab:tabular_analysis}. Numerous datasets, including Newsroom~\citep{newsroom}, CNN Daily Mail~\citep{sarvanidigital}, CNN Corpus~\citep{cnncorpus}, and CCSum~\citep{jiang-dreyer-2024-ccsum}, have been developed for summarization in English. While some of these datasets are publicly accessible, others are privately listed, as listed in Table \ref{tab:tabular_analysis}. Headline generation, as a specific subset of abstractive summarization, has also benefited from datasets, such as XSum~\citep{xsum-dataset}, NewSHead~\citep{representativeheadlines}, PENS~\citep{pens}, New York Times~\citep{AB2/GZC6PL_2008}, DUC~\citep{duc}, and Gigaword~\citep{english-gigaword,gigaword}, all designed for English. Most of these datasets utilize textual content to produce summaries or headlines, with two datasets, NewSHead~\citep{representativeheadlines} and PENS~\citep{pens}, explicitly incorporating category as a feature. While the majority of these works emphasize the abstractive approach, some also explore the extractive methodology.

Beyond English, several datasets have been developed for abstractive headline generation in other languages. These include SuDer~\citep{suder} in Turkish, AFRIHG~\citep{ogunremi2024afrihg} in African, RIA~\citep{gavrilov2019self} and Lenta~\citep{lenta} in Russian, and Mukhyansh~\citep{madasu-etal-2023-mukhyansh} and Varta~\citep{aralikatte-etal-2023-varta} in Indic languages. Additionally, LCSTS~\citep{hu-etal-2015-lcsts} has been created in Chinese for abstractive summarization. Notably, for non-English datasets, the focus has also primarily been on utilizing content alone for headline generation tasks.

In the Bengali language, several datasets have been developed for summarization and headline generation tasks. XL-Sum~\citep{hasan2021xlsumlargescalemultilingualabstractive}, a multilingual dataset covering 44 languages, includes Bengali and focuses on abstractive summarization using news content. Potrika~\citep{potrika} stands out as a substantial dataset offering a large collection of Bengali news samples with different categories, excluding religious news. BNAD~\citep{SAAD2024110874} is another notable dataset in Bengali, which integrates category information with news articles for headline generation, though it includes only 1,275 religious news samples across its domains. Shironaam~\citep{akash-etal-2023-shironaam} is a large-scale dataset encompassing 13 news domains, including religious news, albeit with limited samples in this domain.

Among the previously mentioned research efforts, none of the non-Bengali datasets include the religious news domain. While some Bengali datasets, such as BNAD~\citep{SAAD2024110874} and Shironaam~\citep{akash-etal-2023-shironaam}, incorporate religious news along with categories, they lack subdivisions within the religious content, aspect, and sentiment of a news article. In our proposed dataset, \textit{BeliN}, we address this gap by focusing on the less explored religious news domain. We have compiled religious news samples from various online newspapers, integrating additional features such as category, aspect, and sentiment. The dataset includes five religious categories: Islam, Hinduism, Christianity, Buddhism, and others. Furthermore, it captures four distinct aspects: religious reports, festivals, education, and culture. Additionally, we annotate the sentiment polarity of the news articles as positive, negative, or neutral to further enhance the dataset's utility. 

Numerous headline-generation systems have been developed utilizing various datasets, with the majority relying solely on content as input for generating headlines~\citep{putyourvoice, turkisnews, ogunremi2024afrihg, bukhtiyarov, jin2020hooksheadlinelearninggenerate, representativeheadlines, gigaword, userengagingheadline, singh2021sheg, 9507422, AB2/GZC6PL_2008}. This content-only approach has also been widely adopted in Bengali, as seen in research on text summarization~\citep{hasan2021xlsumlargescalemultilingualabstractive}, and headline generation~\citep{salehin2019generating} that utilize custom datasets. In the news content-only approach, the absence of linguistic context (i.e., the surrounding language or text that provides clarification and deeper meaning beyond the literal interpretation of words~\citep{theledi2024president}) along with the lack of additional guidance, often leads to limited headline diversity and challenges in evaluation due to reliance on a single ground truth~\citep{liu-etal-2020-diverse}. To address this limitation, incorporating additional contextual features is crucial for generating more nuanced and linguistically informed headlines. While Shironaam~\citep{akash-etal-2023-shironaam} introduced a multi-input framework by incorporating category and image captions alongside content, our proposed \textit{MultiGen} approach goes a step further by integrating contextual features such as aspect and sentiment, in addition to content and category. By leveraging sentiment to reflect the natural tone of the news, our method aims to generate more robust and contextually accurate headlines. The \textit{MultiGen} approach demonstrates superior performance compared to the baseline content-only systems, highlighting its effectiveness over existing methods like Shironaam~\citep{akash-etal-2023-shironaam}.

In summary, while headline generation has seen substantial progress in high-resource languages, research in low-resource languages like Bengali remains underexplored. This study bridges the gap by introducing \textit{BeliN}, a specialized dataset for Bengali religious news, and \textit{MultiGen}, a state-of-the-art approach tailored for this domain. Together, they lay the groundwork for broader advancements in low-resource language processing and headline generation.

\section{The BeliN Corpus}
\label{sec:belin}
\begin{figure}[!htb]
  \centering
  \includegraphics[width=1\textwidth]{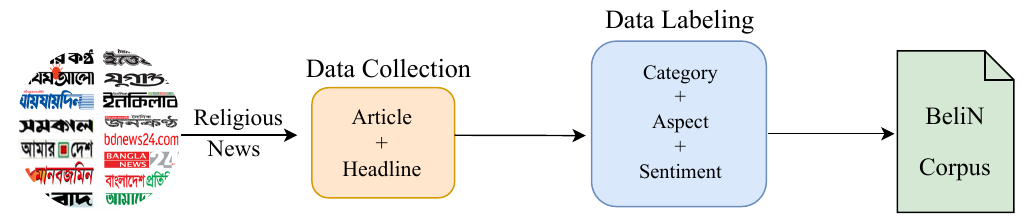}
  \caption{An overview of \textit{BeliN} corpus development process.}
  \label{fig:dataset}
\end{figure}

This section provides a comprehensive overview of the \textit{BeliN} corpus\footnote{\url{https://github.com/akabircs/BeliN}}, a meticulously curated dataset designed to advance the task of Bengali news headline generation. The corpus development process, as illustrated in Figure \ref{fig:dataset}, follows a structured approach that integrates raw data collection, labeling, and statistical analysis to ensure a high-quality and contextually enriched dataset. The process begins with sourcing raw data from diverse Bengali news websites and religious news portals to achieve a representative dataset. This is followed by a detailed labeling methodology, where additional metadata such as categories, sentiments, and aspects are assigned to enrich the data. These auxiliary features enhance the context for generating precise and meaningful headlines. The final stage involves a thorough statistical analysis to evaluate the dataset’s composition and suitability for training and evaluating generative models. Each of these steps is systematically discussed in the following subsections

\subsection{Raw Data Collection}
The raw data collection for the \textit{BeliN} corpus was conducted with the objective of creating a robust dataset for Bengali news headline generation, specifically targeting religious news. Articles and corresponding headlines were manually gathered from a diverse set of Bengali news websites and religious news portals, ensuring the inclusion of high-quality and contextually relevant data. The sources for these articles are listed in Table \ref{tab:newswebsite}, reflecting a wide coverage of topics and perspectives within the religious domain.

The manual collection approach was crucial in ensuring the integrity of the data, particularly in capturing nuanced contexts and maintaining linguistic authenticity. Unlike automated scraping techniques, this method allowed for the careful selection of articles and headlines that align with the focus of the BeliN corpus. This step laid the groundwork for the subsequent labeling process, where additional metadata was assigned to further enrich the dataset’s contextual depth.
\begin{table}[!ht]
    \centering
    \caption{List of newspapers}  \label{tab:newswebsite}
    \begin{tabular}{ll} 
    \hline
       Newspaper & URL \\ \hline\hline
         Prothom alo  & \url{https://www.prothomalo.com/religion} \\ 
         Kaler kantho  & \url{https://www.kalerkantho.com/online/Islamic-lifestylie}  \\ 
         Bangladesh pratidin  & \url{https://www.bd-pratidin.com/islam} \\

         NayaDiganta  & \url{https://www.dailynayadiganta.com/diganta-islami-jobon/133}  \\ 

         Jugantor & \url{https://www.jugantor.com/all-news/islam-life}  \\ 

        Daily Ittefaq  & \url{https://www.ittefaq.com.bd/religion}  \\

         Samakal  & \url{https://samakal.com/search?q=religion}  \\ 

         Dhaka Tribune  & \url{https://www.dhakatribune.com/topic/religion}  \\ 

         Bhorer Kagoj  & \url{https://www.bhorerkagoj.com/religion}  \\ 

         Jai Jai Din  & \url{https://www.jaijaidinbd.com/islam-and-religion}  \\ 

         Alokito Bnagladesh  & \url{https://www.alokitoBengalidesh.com/islam}  \\ 
         Daily Inqilab  & \url{https://dailyinqilab.com/islamic-world}  \\ 
        Daily Vorer Pata  & \url{https://www.dailyvorerpata.com/cat.php?cd=293}  \\ 
         Daily Khabar Patra  & \url{https://khoborpatrabd.com/?s=religion}  \\ 
        \hline

        \hline
    \end{tabular} 
\end{table}
\subsection{Dataset Labeling}
Each article in the BeliN corpus was manually labeled for accuracy and consistency, categorized by religious affiliation, aspect, and sentiment. A subset was further annotated with predefined aspects and corresponding sentiment polarity to aid headline generation. The dataset features a diverse collection of religious news articles from various Bengali sources, structured into five key columns, detailed as follows:

\begin{enumerate}
    \item Article: The full text of the news article.
    \item Headline: The original headline of the news article.
    \item Category: The religious affiliation of the news article -  Islam, Hinduism, Christianity, Buddhism, Others.  
    \item Aspect: This identifies the specific focus or theme of the article's content, which can include categories such as religious reports, festivals, education, culture, or other related topics which can be one of the following:
    \begin{enumerate}
        \item Religious Report: Religious reports typically encompass a range of religious discussions, including sacred events, and mythological and religious tales. These reports highlight significant news about religious communities, broader spiritual perspectives, and important religious philosophies. For example, ``In Pakistan, a church was vandalized and set on fire. Two members of the community were arrested for blasphemy\footnote{\url{https://www.prothomalo.com/world/pakistan/i3riizd976}}".
        
        \item Religious Festival: This section publishes news about religious festivals, ceremonies, and rituals. It includes significant news about various religious communities' festivals, worship, and other religious practices. During festival times, news about events of different religious communities may also be featured in this section. For example, ``Durga Puja is celebrated with grandeur in Abu Dhabi\footnote{\url{https://www.bd-pratidin.com/probash-potro/2023/10/22/932700}}".
        
        \item Religious Education: This section focuses on news related to religious education and spiritual growth. It highlights updates from various educational institutions, religious schools, and policies on religious education. For example, ``Those deeds by which one can attain paradise with the Prophet Muhammad (peace be upon him)\footnote{\url{https://www.kalerkantho.com/online/Islamic-lifestylie/2023/10/25/1330109}}".
        
        \item Religious Culture: In this section, notable news about religious culture and religious characters are mentioned. Religious personalities, mythologies, and religious-cultural events can be particularly highlighted here. For example, ``A 10-day Islamic book fair begins in Mymensingh\footnote{\url{https://www.kalerkantho.com/online/Islamic-lifestylie/2023/10/05/1324183}}".
    \end{enumerate}
    
    \item Sentiment: This represents the sentiment of the article, which can be classified as:
    \begin{enumerate}
        \item Positive: This sentiment label indicates that the content of the article expresses favorable, supportive, or optimistic views toward the subject matter. For example, ``Faith grows through contemplation and research\footnote{\url{https://www.kalerkantho.com/online/Islamic-lifestylie/2023/10/06/1324389}}".

        \item Negative: This sentiment label indicates that the content of the article conveys unfavorable, critical, or pessimistic views toward the subject matter. For example ``The national wealth is at risk of self-destruction\footnote{\url{https://www.kalerkantho.com/online/Islamic-lifestylie/2023/09/02/1314294}}".
        
        \item Neutral: This sentiment label indicates that the content of the article maintains an impartial, balanced, or indifferent stance, without showing strong positive or negative opinions. For example, ``The rare coin of the Islamic era in Saudi Arabia\footnote{\url{https://www.kalerkantho.com/online/Islamic-lifestylie/2023/09/10/1316757}}".

    \end{enumerate}
\end{enumerate}
A sample of the dataset has been given in Table \ref{tab:sample_of_dataset}. The BeliN corpus captures the complexity of religious news, reflecting diverse aspects and sentiments within the articles. 
\begin{table}[!ht]
\caption{Samples of the \textbf{BeliN} corpus}
\includegraphics[width=\textwidth]{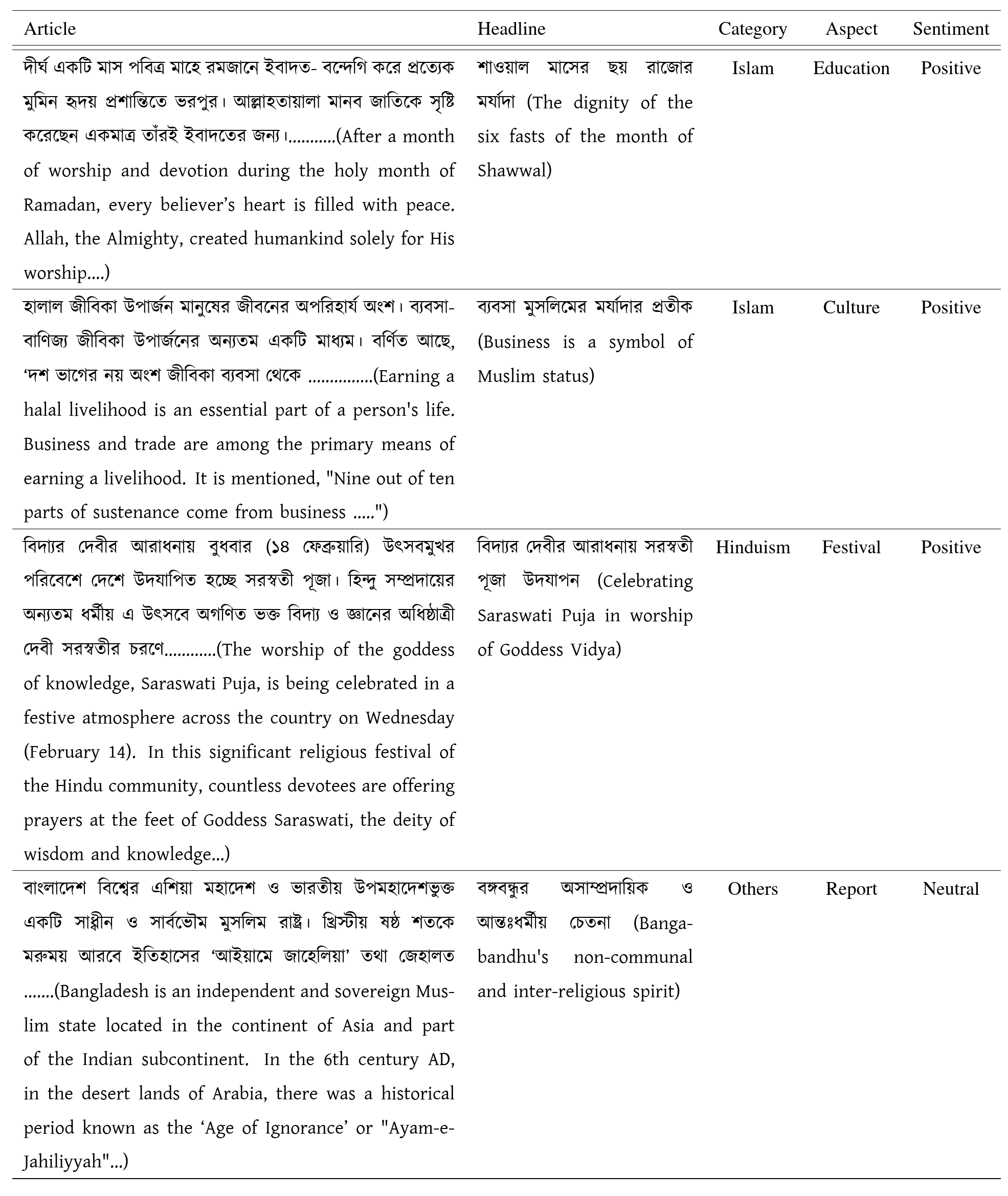}
\label{tab:sample_of_dataset}
\end{table}
\subsection{Dataset Statistics}
This subsection provides a detailed statistical analysis of the BeliN corpus, highlighting its composition and diversity. The dataset, specifically curated for religious news, spans multiple categories, aspects, and sentiment polarities. Such granularity ensures the dataset's utility for training and evaluating generative models, offering a rich context for generating headlines. Table \ref{tab:dataset_stats} presents the statistics of the BeliN corpus, which encompasses religious news across different categories. It includes counts for five major categories. Each category is further analyzed based on four aspects, and sentiment counts (positive, negative, and neutral) are provided for each aspect. The table concludes with a total count of 2520 entries in the dataset, with individual category counts distributed accordingly.
\begin{table}[!ht]
\caption{Descriptive statistics of the \textit{BeliN} corpus}
\setlength{\arrayrulewidth}{0.3pt}
\renewcommand{\arraystretch}{0.7} 
\resizebox{1\textwidth}{!}{%
\begin{tabular}{@{\extracolsep{4pt}}lrrrrrrrr}
\hline

\multirow{2}{*}{Category} & \multicolumn{4}{c}{Aspect} & \multicolumn{3}{c}{Sentiment} & \multirow{2}{*}{Total} \\
\cline{2-5}\cline{6-8}
 &   Report &  Festival &  Education &  Culture & Positive & Negative & Neutral &\\
\hline\hline
 Islam & 860 & 68 & 890 & 183 & 1457 & 299 & 245  & 2001  \\

Hinduism  & 135 & 67 & 16 & 24 & 128 & 58 & 56 & 242  \\

 Christianity  & 7 & 12 & 7 & 2 & 19 & 5 & 4 & 28 \\

Buddhism  & 12 & 13 & 1 & 3 & 25 & 3 & 1 & 29   \\

Others  & 190 & 1 & 16 & 13 & 88 & 90 & 42 & 220   \\
\hline
Total & 1204 & 161 & 930 & 225 & 1717 & 455 & 348 & 2520\\
\hline

\hline
\end{tabular}
}
\label{tab:dataset_stats}
\end{table}

Table \ref{tab:features} compares the features of Shironaam and BeliN in the religious domain. BeliN includes additional features like aspect and sentiment, while Shironaam includes topic words and image captions. BeliN also has a larger number of samples (2520 news) compared to Shironaam (294 news). 
\begin{table}[!ht]
\centering
\caption{Feature comparison of the \textit{BeliN} dataset and Shironaam.}
\begin{tabular}{lcc}
\hline
Features          & Shironaam~\citep{akash-etal-2023-shironaam} & BeliN (this study) \\ 
\hline
\hline
Article                    & \checkmark           & \checkmark               \\ 
Headline                   & \checkmark           & \checkmark               \\ 
Category                   & \checkmark          & \checkmark              \\ 
Aspect                & \texttimes           & \checkmark               \\ 
Sentiment              & \texttimes           & \checkmark               \\ 
Topic words                & \checkmark           & \texttimes               \\ 
Image caption              & \checkmark           & \texttimes               \\ 
Total Samples                 & 294*              & 2520                  \\ \hline

\hline
\multicolumn{3}{l}{* Religious news}
\end{tabular}
\label{tab:features}
\end{table}

\begin{figure}[!ht]
\centering
\begin{subfigure}{1\textwidth}
    \centering
    \includegraphics[width=1\textwidth]{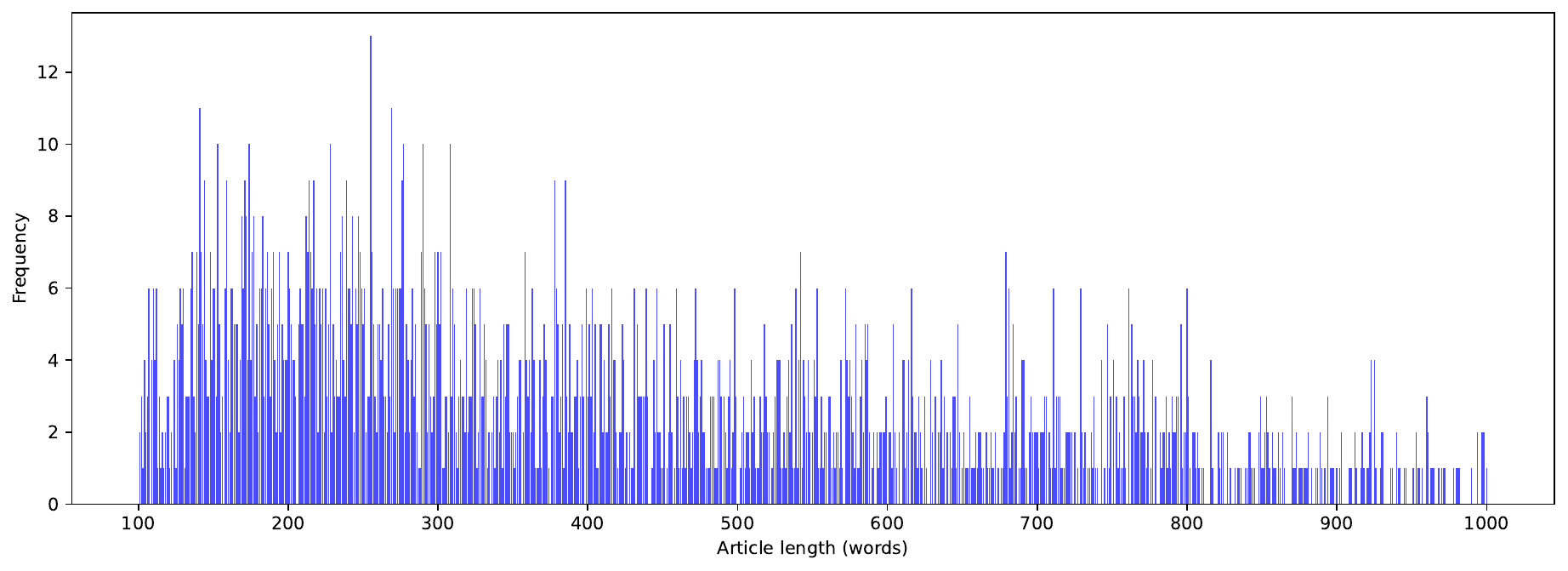}  
    \caption{}
    \label{fig:farticle}
\end{subfigure}
\begin{subfigure}{1\textwidth}
    \centering
    \includegraphics[width=1\textwidth]{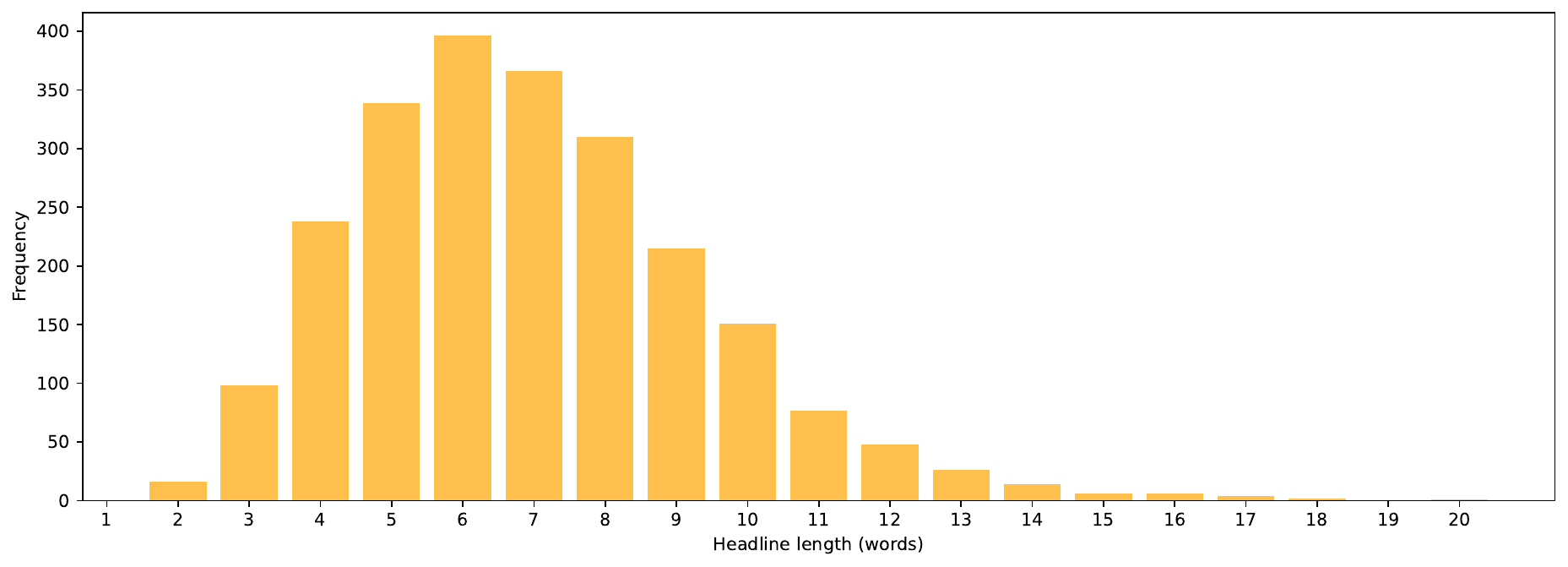} 
    \caption{}
    \label{fig:fheadline}
\end{subfigure}
\caption{Distribution of article and headline lengths in the dataset. (a) Frequency of article lengths (in words), illustrating the common word counts for articles. (b) Frequency of headline lengths (in words), highlighting the typical brevity or elaboration of headlines compared to the full articles.}
\label{fig:fdist}
\end{figure}

Table \ref{tab:quantt} shows the quantitative statistics of Shironaam and BeliN based on the average number of words, sentences, and vocabulary size. The BeliN dataset demonstrates strong novelty in its n-grams, with 4.42\% novel unigrams, 21.48\% novel bigrams, 42.10\% novel trigrams, and 56.47\% novel 4-grams, as shown in Table~\ref{tab:novel_n_grams}. These results highlight the distinctiveness of headlines in comparison to the articles. While Shironaam shows slightly higher percentages of novel n-grams, BeliN still provides valuable insights into Bengali religious news, showcasing significant diversity in language use. This makes BeliN a valuable resource for research in Bengali language processing. The figures presented illustrate the distribution of article and headline lengths in the dataset. Figure~\ref{fig:farticle} shows the frequency of article lengths, measured in words, revealing the common word counts for articles. Figure~\ref{fig:fheadline} depicts the frequency of headline lengths, also measured in words, providing insight into the typical brevity or elaboration of headlines compared to the full articles. Together, these figures offer a visual representation of the structure and variation in article and headline lengths within the dataset.
\begin{table}[!ht]
    \centering
    \caption{Comparison of the \textit{BeliN} and Shironaam datasets based on average words, average sentences, and vocabulary size.} \label{tab:quantt}
    \resizebox{1\textwidth}{!}{
    \begin{tabular}{@{\extracolsep{4pt}}crrrrrr}
        \hline
        \multirow{2}{*}{Dataset} & \multicolumn{3}{c}{Article} & \multicolumn{3}{c}{Headline} \\ \cline{2-4}\cline{5-7} 
        & Avg. words & Avg. sentences & Vocabulary & Avg. words & Avg. sentences & Vocabulary \\ \hline\hline
        Shironaam*~\citep{akash-etal-2023-shironaam} & 943.43 & 7.55 & 3,497 & 13.03 & 1.02 & 416 \\
        BeliN (this study)    & 1001.18 & 32.75 & 9,750 & 17.13 & 1.06 & 1,410 \\ 
        \hline

        \hline
        \multicolumn{7}{l}{* for religious news only}
    \end{tabular}
    }
   
\end{table}

\begin{table}[!ht]
\centering
\caption{Percentage of n-grams in the \textit{BeliN} and Shironaam datasets.}
\begin{tabular}{lccccc}
\hline

Dataset & Unigram & Bigram & Trigram & 4-gram \\\hline\hline
Shironaam*~\citep{akash-etal-2023-shironaam} & 4.50\% & 22.48\% & 45.19\% & 60.22\% \\
BeliN (this study) & 4.42\% & 21.48\% & 42.10\% & 56.47\% \\
\hline

\hline
 \multicolumn{4}{l}{* for religious news only}
\end{tabular}
\label{tab:novel_n_grams}
\end{table}

The rich contextual information embedded in the BeliN corpus has significant potential for various natural language processing (NLP) tasks  beyond headline generation. These include text generation, topic modeling, news categorization, news headline sentiment analysis within the realm of religious news. By leveraging the detailed annotations and multi-aspect nature of the dataset, researchers and developers can create more sophisticated and human-like AI systems capable of understanding and generating content with high contextual awareness.

\section{The \textit{MultiGen} Approach}
\label{sec:multigen}
The traditional news content-only approach to headline generation faces several challenges. Relying solely on the news content often results in a lack of linguistic context and guidance, limiting headline diversity and creating evaluation difficulties due to dependence on a single ground-truth reference~\citep{liu-etal-2020-diverse}. In this approach, the model is designed to take solely news content as input to generate headlines, which may not capture the full spectrum of possible interpretations or nuances of the article. Additionally, without the inclusion of contextual features, generated headlines may fail to align with the emotional tone or thematic aspects of the content, diminishing their relevance and overall quality.

To overcome these limitations, the \textit{MultiGen} approach introduces a multi-input framework for Bengali news headline generation. Unlike the conventional approach that relies exclusively on news content, \textit{MultiGen} integrates additional contextual features such as aspect, category, and sentiment. These features provide a more comprehensive understanding of the article, allowing the model to generate headlines that are contextually relevant and linguistically informed. Aspect and category help the model focus on specific narrative elements, while sentiment captures the emotional tone, ensuring that the generated headlines are better aligned with the article's mood and message. This enriched approach improves both the diversity and quality of the generated headlines. Incorporating contextual features has also proven effective in other similar NLP tasks, including text classification~\citep{kiefer2022case}, information retrieval~\citep{chen2019ci}, and sentiment analysis~\citep{zhu2023multimodal,aziz2023mmtf}.

\begin{figure}[!ht]
    \centering
    \includegraphics[width=\textwidth]{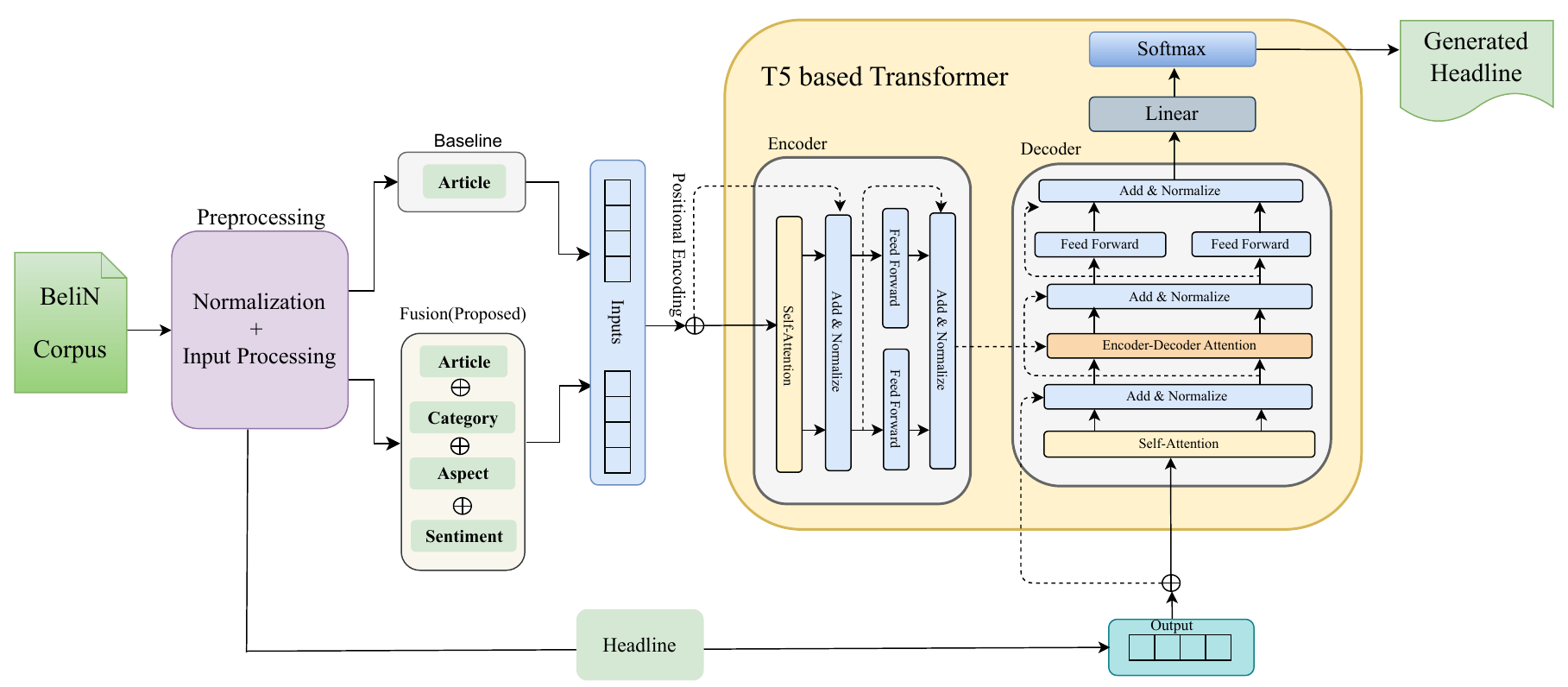}
    \caption{\textit{MultiGen} architecture for headline generation}
    \label{fig:t5}
\end{figure}

By incorporating these additional features, \textit{MultiGen} enhances the model's ability to understand the nuances of the article, producing headlines that are both informative and contextually aligned with the content's emotional tone. The overall framework of \textit{MultiGen}, as illustrated in Figure \ref{fig:t5}, showcases the seamless integration of these diverse inputs within an encoder-decoder architecture, enabling more accurate and contextually aware headline generation.
The remainder of this section provides a detailed description of the \textit{MultiGen} approach, starting with the preprocessing steps, followed by the fusion of multiple inputs, and concluding with the encoder-decoder architecture used to achieve enhanced headline generation performance.

\subsection{Preprocessing}  
Preprocessing is a critical phase that prepares raw text data for input into generative models, ensuring the data is clean, consistent, and ready for effective processing. It involves two main tasks:  
\begin{itemize}
    \item Text Normalization: This step uses the BUET normalizer \citep{hasan-etal-2020-low} to standardize characters with Unicode NFKC. Non-textual elements like URLs and emojis are removed, excessive whitespace is managed, and redundant punctuation characters are reduced. 
     \item Input Processing: Text is formatted for model training by adding task-specific prefixes, such as ``Summarize the Article as Headlines,'' to guide the model. Appropriate tokenizers, like AutoTokenizer for BanglaT5 and mBART, are used to process articles. The text is truncated to 512 tokens for input and 64 tokens for headlines, ensuring computational efficiency. Finally, tokenized inputs and labels are combined into a dictionary for training, enhancing coherence and output quality.  
\end{itemize} 

\subsection{Fusing Article with Category, Aspect, and Sentiment}  The proposed multi-input approach enhances Bengali news headline generation by integrating multiple contextual signals—article (\(\mathbf{A}\)), category (\(\mathbf{C}\)), aspect (\(\mathbf{P}\)), and sentiment (\(\mathbf{S}\))—into a unified input sequence. By leveraging this enriched input, the model benefits from a broader contextual understanding, enabling it to generate more precise and contextually aligned headlines. This approach constructs the input sequence by fusing these components with a [SEP] token, ensuring a clear distinction between different elements while preserving their individual contributions. In contrast to the baseline approach, which relies solely on article content, the proposed fusion approach incorporates additional contextual elements, significantly enhancing the model’s capacity to produce contextually nuanced headlines. Specifically, the input sequence \(\mathbf{I}\) is defined as:  

\[
\mathbf{I} = [A_1, \ldots, A_n, \text{[SEP]}, C_1, \ldots, C_k, \text{[SEP]}, P_1, \ldots, P_m, \text{[SEP]}, S_1, \ldots, S_p]
\]

This fusion strategy, referred to as \textit{MultiGen}, is designed to enhance the quality of headline generation by tailoring the outputs to specific categories, aspects, and sentiment polarities. By capturing richer contextual information, this approach generates headlines that resonate more closely with the article’s intent and emotional tone, ultimately delivering a more engaging and context-aware reader experience.

\subsection{Encoder-Decoder Architecture}
The T5 (Text-to-Text Transfer Transformer) model employs an encoder-decoder architecture specifically designed for sequence-to-sequence tasks. This architecture consists of two primary components: the encoder and the decoder, both built using transformer layers. The encoder processes the input sequence and converts it into fixed-length continuous hidden representations (\(\mathbf{Z}\)) that encapsulate both semantic and syntactic features. The decoder then utilizes these hidden representations to generate the output sequence by predicting tokens iteratively, conditioned on the encoder's output and previously generated tokens.

\paragraph{Encoder} 
The encoder comprises a stack of transformer layers, each including multi-head self-attention mechanisms and feedforward neural networks. Given an input sequence, the encoder first embeds the tokens into continuous vector representations and applies positional encodings to capture the order of tokens. Successive transformer layers refine these representations by attending to various parts of the input, capturing both local and global dependencies. Mathematically, the encoder maps the input sequence (\(\mathbf{I}\)) into hidden states (\(\mathbf{Z}\)) as follows:
\[
\mathbf{Z} = f_{\theta_{\text{enc}}}(\mathbf{I})
\]

\paragraph{Decoder} 
The decoder also consists of transformer layers and incorporates a cross-attention mechanism to attend to the encoder's output representations. During generation, the decoder predicts each token sequentially, leveraging the hidden states from the encoder and previously generated tokens. The output probabilities for the next token are computed using a softmax layer over the vocabulary. Formally, the decoder generates the output sequence (\(\mathbf{H}\)) conditioned on the encoder's hidden states (\(\mathbf{Z}\)):
\[
p_{\theta_{\text{dec}}}(\mathbf{H} \mid \mathbf{Z}) = p_{\theta_{\text{dec}}}(\mathbf{H} \mid f_{\theta_{\text{enc}}}(\mathbf{I}))
\]
In the baseline approach, the encoder processes only the news content (\(\mathbf{A}\)) to generate hidden states, which the decoder uses to produce the headline (\(\mathbf{H}\)):
\[
\mathbf{Z} = f_{\theta_{\text{enc}}}(\mathbf{A})
\]
\[
p_{\theta_{\text{dec}}}(\mathbf{H} \mid \mathbf{Z}) = p_{\theta_{\text{dec}}}(\mathbf{H} \mid f_{\theta_{\text{enc}}}(\mathbf{A}))
\]

In the proposed approach, the input sequence (\(\mathbf{I}\)) includes additional contextual information—category (\(\mathbf{C}\)), aspect (\(\mathbf{P}\)), and sentiment (\(\mathbf{S}\))—concatenated with the news content using [SEP] tokens:
\[
\mathbf{I} = [A_1, \ldots, A_n, C_1, \ldots, C_k, P_1, \ldots, P_m, S_1, \ldots, S_p]
\]
The encoder transforms this enriched input into hidden representations (\(\mathbf{Z}\)), which the decoder leverages to generate contextually enriched headlines:
\[
\mathbf{Z} = f_{\theta_{\text{enc}}}(\mathbf{I})
\]
\[
p_{\theta_{\text{dec}}}(\mathbf{H} \mid \mathbf{Z}) = p_{\theta_{\text{dec}}}(\mathbf{H} \mid f_{\theta_{\text{enc}}}(\mathbf{I}))
\]

This architecture's ability to incorporate auxiliary information into the input sequence enables it to generate more accurate and contextually relevant headlines, demonstrating the effectiveness of the proposed enhancements.

\section{Experimental Evaluation}
\label{sec:evaluation}
\subsection{Experimental Settings}
For our Bengali religious news headline generation task, we employed a combination of hardware and software resources to ensure efficient model training and evaluation. Hardware resources included Google Colab Pro with an NVIDIA A100 GPU, Kaggle's environment with NVIDIA T4 $\times$ 2 GPUs, and a local machine equipped with an NVIDIA T4 GPU to maximize computational efficiency. On the software side, the project was developed on a Windows 11 machine with a 1TB HDD and 512GB SSD. We utilized TensorFlow and PyTorch for deep learning, NLTK for text preprocessing, and Hugging Face's Transformers library for implementing encoder-decoder Transformer architectures. The dataset was split into training (1870 samples, 74\%), validation (150 samples, 6\%), and testing (500 samples, 20\%) subsets, allowing for thorough model training, hyperparameter tuning, and performance evaluation. This setup provided a solid foundation for the successful development and assessment of the headline generation models.

\subsection{Evaluation Metrics}
To evaluate the performance of our developed system, we utilized several evaluation metrics. These metrics include BLEU, ROUGE-1, ROUGE-2, ROUGE-L, BERTScore and  METEOR.

\paragraph{ROUGE (Recall-Oriented Understudy for Gisting Evaluation)} This metric is a widely used family of metrics for evaluating natural language processing tasks, particularly text summarization \citep{lin2004rouge}. It measures the similarity between model-generated summaries and reference summaries based on n-gram overlap and the longest common subsequence (LCS). Three key ROUGE metrics are commonly employed: ROUGE-1 evaluates the overlap of unigrams (individual words) between the generated and reference summaries, calculating Precision and Recall. ROUGE-2 extends this to bi-grams (pairs of consecutive words), measuring the degree of structural similarity between the summaries. ROUGE-L, on the other hand, focuses on the longest common subsequence (LCS), which captures the longest sequence of words that appears in both summaries, regardless of order, providing a measure of structural alignment. Together, these ROUGE metrics offer a comprehensive view of content and structural similarity between generated and reference summaries, helping assess the effectiveness of the summarization model.

\paragraph{BLEU (Bilingual Evaluation Understudy)} BLEU~\citep{papineni2002bleu} is a widely used metric for evaluating the quality of machine-translated text. It measures the similarity between model-generated and human-generated reference translations based on n-gram precision. BLEU is calculated by comparing the n-grams (sequences of n words) generated by the model to the n-grams in the reference translations. The BLEU score ranges from 0 to 1, with higher scores indicating better translation quality. The BLEU score can be expressed mathematically as:  
        \begin{equation}
            \text{BLEU} = \text{BP} \times \exp \left( \sum_{n=1}^{N} \frac{1}{N} \log \left( \text{precision}_n \right) \right)
        \end{equation}
where $\text{BP}$ is the brevity penalty to account for shorter translations, and $\text{precision}_n$ is the modified precision for n-grams of size $n$. The brevity penalty $\text{BP}$ is calculated as:
        \begin{equation}
        \text{BP} = 
        \begin{cases} 
        1 & \text{if } c > r \\ 
        \exp(1 - \frac{r}{c}) & \text{if } c \leq r 
        \end{cases}
        \end{equation}
where $c$ is the length of the model output and $r$ is the effective reference length, which is the length of the reference translation closest to the length of the model output. BLEU is a valuable metric for evaluating the overall quality of machine-generated translations, but it has some limitations, particularly in capturing semantic similarity and fluency. It is best used in conjunction with other evaluation metrics for a comprehensive assessment of translation quality.

\paragraph{METEOR (Metric for Evaluation of Translation with Explicit ORdering)} This metric \citep{banerjee2005meteor} is designed to evaluate machine translation by considering synonyms, stemming, and paraphrasing. It calculates precision and recall based on alignments between the generated and reference texts. The METEOR score is the harmonic mean of precision and recall, adjusted by a penalty for fragmentation. Mathematically, the METEOR score is given by:
            \begin{equation}
            F_{\text{mean}} = \frac{10 \cdot P \cdot R}{R + 9 \cdot P}
            \end{equation}
            \begin{equation}
            \text{METEOR} = F_{\text{mean}} \times (1 - \text{penalty})
            \end{equation}
            
\paragraph{BERTScore} It evaluates the quality of text generation using contextual embeddings \citep{bertscore} from pre-trained BERT models. It calculates the cosine similarity between token embeddings of the generated and reference texts. BERTScore includes precision, recall, and F1-score based on these similarities. Mathematically, BERTScore is defined as:
            \begin{equation}
            \text{Precision} = \frac{1}{|X|} \sum_{x \in X} \max_{y \in Y} \text{cosine}(x, y)
            \end{equation}
            \begin{equation}
            \text{Recall} = \frac{1}{|Y|} \sum_{y \in Y} \max_{x \in X} \text{cosine}(y, x)
            \end{equation}
            \begin{equation}
            \text{F1-score} = 2 \times \frac{\text{Precision} \times \text{Recall}}{\text{Precision} + \text{Recall}}
            \end{equation}
where \(X\) and \(Y\) represent the sets of token embeddings for the model output and reference translations, respectively.

\subsection{Pre-trained Language Models}
The pre-trained language models employed for this task include both T5-based and BART-based models. These models are fine-tuned to transform Bengali news articles into concise, informative headlines. The T5-based models include BanglaT5, mT5, and mT0, while mBART represents the BART-based model. Detailed information about these models is provided in Table \ref{tab:models}.

\begin{itemize}

\item T5-based models: BanglaT5, mT5, and mT0 are based on the T5 architecture \citep{JMLR:v21:20-074}, which frames all NLP tasks as text-to-text problems. The T5 model uses a sequence-to-sequence framework where the encoder processes the input text and the decoder generates the output text. This approach allows for flexibility in handling various NLP tasks, such as translation, summarization, and text generation, by converting them into a text-to-text format.

\item BART-based model: mBART is based on the BART architecture \citep{lewis2019bart}, which utilizes a denoising autoencoder approach for pre-training. This model follows a sequence-to-sequence framework similar to T5 but incorporates a denoising objective during pre-training, where parts of the input are corrupted and the model learns to reconstruct the original text. This pre-training strategy helps the model become robust to noise and enhances its ability to generate coherent and contextually relevant text. 
\end{itemize}

\begin{table}[!hbt]
    \centering
    \caption{Details of the pre-trained language models used.}
    \begin{tabular}{l l c  c } \hline
       Model & Hugging Face link & Parameters & Pretrained on \\  \hline \hline
        Bangala-T5 \citep{bhattacharjee2022banglanlg} & \url{https://huggingface.co/csebuetnlp/BanglaT5} & 247M & Bengali2B+ \\ 
        mT0-base \citep{muennighoff2022crosslingual} & \url{https://huggingface.co/bigscience/mt0-base} & 582M & mC4 \\ 
        mT5-Base \citep{xue2021mt5} & \url{https://huggingface.co/google/mt5-base} & 582M & mC4 \\ 
        mBART-50 \citep{tang2020multilingual} & \url{https://huggingface.co/facebook/mbart-large-50} & 610M & CC25 \\ \hline 
        
        \hline
    \end{tabular}
    \label{tab:models}
\end{table}
\subsection{Hyper-parameter Tuning}  
Hyperparameter tuning is a critical step in optimizing the performance of generative models for headline generation. In this study, we experimented with various hyperparameter configurations for Bangla-T5, mBART, mT5, and mT0 to achieve the best results. Table~\ref{tab:hyperparams} summarizes the hyperparameter search space and the selected configurations for each model.

\begin{table*}[!hbt]
    \centering
    \caption{The hyperparameter search space used in tuning and the selected optimal hyperparameters for each model.}
    \begin{tabular}{ l ccccc  } \hline
       Hyper-parameters  &  Hyper-parameter Space & Bangla-T5 & mBART & mT5 & mT0 \\ \hline\hline 
       Learning Rate   & \(2 \times 10^{-5}\), \(1 \times 10^{-4}\), \(1 \times 10^{-3}\) & \(1 \times 10^{-4}\) & \(1 \times 10^{-3}\) & \(1 \times 10^{-4}\) & \(2 \times 10^{-5}\) \\  
       Epochs   & 3--10 & 5 & 5 & 5 & 5 \\ 
       Batch Size   & 4, 8 & 8 & 8 & 8 & 8 \\ 
       Input Token Length   & 512, 1024 & 512 & 512 & 512 & 512 \\ 
       Target Token Length   & 16, 32, 64, 128 & 64 & 64 & 64 & 64 \\ 
       \hline
    \end{tabular}
    \label{tab:hyperparams}
\end{table*}

The selection of appropriate hyperparameters directly impacts the performance and efficiency of the models. The learning rate, for instance, was found to be a key factor in stabilizing training. While Bangla-T5 and mT5 performed optimally with a learning rate of \(1 \times 10^{-4}\), mBART required a higher rate of \(1 \times 10^{-3}\), and mT0 performed best with \(2 \times 10^{-5}\). A batch size of 8 was chosen to balance computational requirements and model convergence, ensuring stable training. Input and target token lengths were also crucial for effectively handling the variability in article lengths and headline requirements. Setting the input token length to 512 ensured that the models could process sufficiently detailed content, while a target token length of 64 allowed the generation of concise and precise headlines without truncation. The number of epochs, fixed at 5, provided a balance between overfitting and undertraining for all models. These carefully chosen hyperparameters enabled the models to achieve strong performance in generating contextually relevant and coherent headlines, underscoring the importance of systematic hyperparameter tuning.

\subsection{Results}
This part evaluates and analyzes the performance of the headline generation models developed in this research. We have provided a comprehensive overview of the evaluation metrics employed, including ROUGE, BLEU, METEOR, and BERTScore. These metrics are widely used in natural language processing tasks to assess the quality of generated text. State of the Art (SOTA) analysis is a critical component of model evaluation, as it establishes a benchmark against which the performance of proposed models can be compared. By analyzing improvements in performance metrics relative to existing methods, researchers can demonstrate the effectiveness and advancements offered by their proposed approaches. The performance of the developed models was evaluated using a combination of these metrics. Table~\ref{table:resize_small_values} reports the results of our models, comparing the proposed approaches against their respective baseline models, and providing insights into the SOTA improvements.
\begin{table}[!ht]
\begin{center}
\caption{Performance comparison of the proposed \textit{MultiGen} approach and the baseline across various transformer-based pre-trained models.}
\begin{tabular}{cccccccc}
\hline
Model & Approach & BLEU & ROUGE-1 & ROUGE-2 & ROUGE-L &BERTScore & METEOR \\

\hline\hline

\multirow{2}{*}{ mT5} &  Baseline &  10.31 &  13.47 &  4.22 &  13.03 &  69.34 &  9.80 \\

 &  Proposed &  {11.66} &  {17.54} &  {5.68} &  {16.85} &  {71.74} &  {10.86} \\
\cline{2-8}
 &  $\Delta$ SOTA &  +13.1\% &  +30.4\% &  +34.6\% &  +29.1\% &  +3.5\% &  +10.8\% \\
\hline

\multirow{2}{*}{ mT0} &  Baseline &  12.08 &  18.84 &  7.10 &  17.95 &  70.34 &  13.90 \\
 &  Proposed &  {13.13} &  {22.94} &  {7.94} &  {21.48} &  {72.62} &  {14.40} \\

\cline{2-8}
 &  $\Delta$ SOTA &  +8.7\% &  +21.2\% &  +11.8\% &  +19.0\% &  +3.2\% &  +3.6\% \\
\hline

\multirow{2}{*}{ mBART} &  Baseline &  15.23 &  23.01 &  7.90 &  21.88 &  73.21 &  13.12 \\

 &  Proposed &  {16.58} &  {24.36} &  {7.78} &  {22.63} &  {74.63} &  {14.60} \\

\cline{2-8}
 &  $\Delta$ SOTA &  +8.8\% &  +5.9\% &  -1.5\% &  +3.4\% &  +1.9\% &  +11.3\% \\
\hline

\multirow{2}{*}{ {BanglaT5}} &  Baseline &  16.08 &  22.84 &  7.97 &  23.08 &  73.57 &  15.40 \\

 &  {Proposed} &  { 18.61} &  { 26.70} &  { 10.60} &  { 24.19} &  { 75.12} &  { 16.65} \\

\cline{2-8}
 &  $\Delta$ SOTA &  +15.7\% &  +17.0\% &  +33.0\% &  +4.8\% &  +2.1\% &  +8.1\% \\
\hline

\hline
\end{tabular}

\label{table:resize_small_values}
\end{center}
\end{table}

The proposed approach consistently outperforms the baseline across all models in terms of BLEU, ROUGE, BERTScore, and METEOR scores. The BLEU scores indicate a notable enhancement in fluency and coherence of the generated headlines, with the proposed models achieving significant improvements over their baselines. For instance, the mT5 model exhibits a BLEU score increase of 13.1\%, while BanglaT5 shows a remarkable improvement of 15.7\%, emphasizing the effectiveness of the proposed methods in producing high-quality outputs. The ROUGE scores further demonstrate the superiority of the proposed approach, revealing higher precision and recall compared to the baseline models. This reflects an improved relevance and informativeness in the generated headlines. For example, mT0's proposed method achieves a ROUGE-1 score that is 21.2\% better than its baseline, indicating that it captures more relevant content. Additionally, BERTScore and METEOR metrics underscore the semantic accuracy and contextual relevance of the generated headlines. The improvements in these scores highlight the effectiveness of the proposed models in understanding and generating text that aligns well with human expectations. Among all the models, BanglaT5 stands out as the top performer in the proposed approach, achieving the highest scores across all metrics. The substantial improvements in BLEU, ROUGE, BERTScore, and METEOR suggest that BanglaT5 effectively leverages additional contextual information, such as aspect categories and sentiment analysis, to generate headlines that are not only accurate and informative but also contextually relevant. Overall, the findings from this research indicate that the proposed models significantly enhance the quality of Bengali news headline generation, offering promising directions for future research and development in this area.

\section{Discussion}
\label{sec:discussion}
\subsection{Analyzing Generated Headlines} 
Table~\ref{tab:sample_dataset} illustrates sample-generated headlines, showcasing the impact of our \textit{MultiGen} approach on generating high-quality, contextually relevant headlines for Bengali news articles. The proposed approach demonstrates significant improvements over the baseline, particularly in preserving the essence of the articles while ensuring linguistic fluency and contextual accuracy. The generated headlines from both the baseline and the proposed \textit{MultiGen} approach were compared against the reference headlines to evaluate their quality and contextual relevance. The analysis reveals notable differences in the performance of the two models, particularly in their ability to align with the reference headlines.

\begin{table}[!ht]
\centering
\caption{Qualitative analysis of headlines generated by the proposed \textit{MultiGen} approach.}
\includegraphics[width=\textwidth]{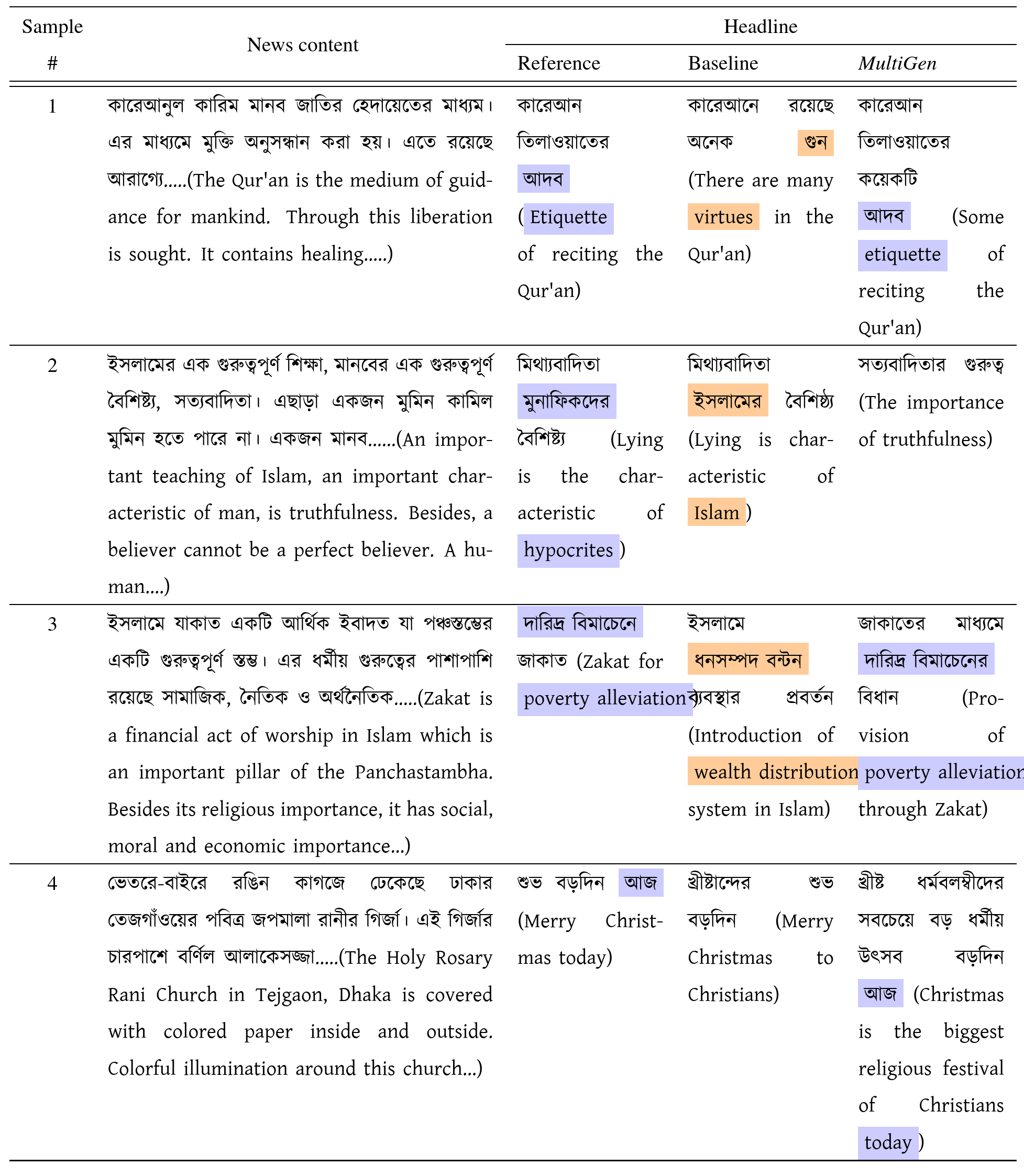}
\label{tab:sample_dataset}
\end{table}

The baseline model, while capable of generating coherent headlines, often fails to capture the nuanced meaning or context of the input article. For instance, in sample \#1, the reference headline focuses on the etiquette of reciting the Qur'an. The baseline model, however, generates a different headline highlighting the virtues of the Qur'an, which does not address the context intended in the reference. In contrast, the proposed \textit{MultiGen} approach successfully generates a headline emphasizing the manners of recitation, aligning more closely with the reference.

Similarly, in sample \#2, the baseline approach generated headline incorrectly attributes lying as a characteristic of Islam, reflecting a lack of semantic understanding. Conversely, the proposed approach accurately conveys the importance of truthfulness, capturing the core message of the article and adhering more closely to the reference.

For sample \#3, the reference headline highlights poverty alleviation through Zakat. While the baseline approach generates a headline related to wealth distribution in Islam, it does not focus on the specific theme of poverty alleviation. The \textit{MultiGen} approach, however, generates a headline that aligns with the reference and integrates the article's broader social and economic implications.

Sample \#4 further illustrates the shortcomings of the baseline approach, which generates a headline that fails to capture the critical detail that today is Christmas Day. In contrast, the proposed approach successfully incorporates this temporal context, producing a headline that accurately conveys the date-related information and aligns closely with the reference.

Overall, the \textit{MultiGen} approach consistently outperforms the baseline model by generating headlines that are contextually and temporally accurate, semantically rich, and closely aligned with the reference. This demonstrates the efficacy of incorporating additional contextual information such as category, aspect, and sentiment, enabling the proposed model to better understand and reflect the underlying themes of the input articles.

\subsection{Findings and Observations}  
This section highlights key insights from the results presented in Table~\ref{table:resize_small_values}, with a focus on error analysis to identify areas where the headline generation models may have performed sub-optimally. Below are the primary observations:

\paragraph{Low BLEU Scores}  
The mT5 model, under the baseline approach, demonstrated relatively low BLEU scores compared to other models. This suggests that the generated headlines often lacked fluency or coherence, resulting in lower n-gram overlaps with the reference headlines.

\paragraph{Variability in ROUGE Scores}  
While the proposed approach generally outperformed the baseline across all models, variability in ROUGE scores was observed. For instance, the ROUGE-2 scores for mT5 and mBART under the proposed approach were slightly lower than other models, indicating difficulties in capturing bi-gram similarities effectively.

\paragraph{Performance Discrepancies}  
BanglaT5 consistently exhibited superior performance, particularly in terms of ROUGE scores, highlighting its ability to generate headlines that closely align with reference headlines. Conversely, mT0's relatively lower BLEU and ROUGE scores suggest room for improvement in fluency and relevance.

\paragraph{Impact of Additional Context}  
The improved performance of the proposed approach, which leverages additional contextual features such as aspect categories and sentiment, underscores the significance of incorporating auxiliary inputs for headline generation. This approach facilitates a deeper understanding of article contexts, leading to more coherent and informative outputs.

\paragraph{Room for Improvement}  
Despite the promising results, there is potential for further improvement in headline generation. Refining model architectures, optimizing hyperparameters, and enhancing preprocessing techniques could mitigate observed shortcomings and elevate the quality of generated headlines.

\subsection{Limitations and Future Work}

Despite the encouraging results, the study encountered several limitations. A significant challenge was the scarcity of high-quality annotated datasets for Bengali, which constrained model training and evaluation, potentially limiting the generalizability of results. Hardware limitations also restricted fine-tuning large-scale generative models, impacting training efficiency and performance. Furthermore, the complexity of Bengali morphology and syntax introduced additional challenges, occasionally resulting in inaccuracies in the generated headlines \citep{rahman2024rise}.

Future work could address these limitations through innovative data augmentation techniques, exploration of advanced model architectures, and domain-specific customization. Expanding the dataset to include a wider variety of domains and incorporating user feedback in training loops could further enhance the models. Additionally, leveraging LLMs for real-time and multilingual headline generation could broaden the applicability and effectiveness of the approach \citep{kabir-etal-2024-benllm}. This study provides a solid foundation for advancing headline generation in Bengali, offering valuable insights for future research in natural language processing and text generation.

\section{Conclusion}
\label{sec:conclusion}
This research work has explored the potential of a contextual multi-input feature fusion approach, using various generative models for religious news headline generation, with a particular focus on the Bengali language. Central to this work is the introduction of the novel \textit{BeliN} corpus, a curated dataset of Bengali religious news articles and corresponding headlines. This dataset addresses the scarcity of resources for Bengali and serves as a foundational contribution to advancing natural language processing for low-resource languages.
We have implemented and evaluated state-of-the-art pre-trained models, including mT5, mT0, mBART, and BanglaT5, within the proposed \textit{MultiGen} approach, incorporating additional contextual information such as aspect, category, and sentiment analysis. Rigorous experimentation and detailed analysis demonstrate that the proposed approach significantly outperforms traditional baseline methods, achieving higher accuracy, coherence, and contextual relevance in generated headlines.

The findings underscore the importance of integrating contextual features in headline generation and highlight the efficacy of the \textit{BeliN} corpus in enabling this advancement. This research contributes to natural language processing and offers practical insights for developing sophisticated text summarization systems in underrepresented languages, thereby promoting linguistic inclusivity and cross-cultural communication.

\section*{Data availability}
Data and code used in this study are publicly available at \url{https://github.com/akabircs/BeliN}.


\section*{Declaration of competing interest}
The authors declare that they have no known competing financial interests or personal relationships that could have appeared to influence the work reported in this paper.

\section*{Acknowledgement}
We want to thank Sadia Tasnim (Begum Rokeya University, Rangpur) for their assistance with collecting and annotating the dataset.

\bibliographystyle{elsarticle-num-names} 
\bibliography{refs}

\begin{thebibliography}{69}
\expandafter\ifx\csname natexlab\endcsname\relax\def\natexlab#1{#1}\fi
\providecommand{\url}[1]{\texttt{#1}}
\providecommand{\href}[2]{#2}
\providecommand{\path}[1]{#1}
\providecommand{\DOIprefix}{doi:}
\providecommand{\ArXivprefix}{arXiv:}
\providecommand{\URLprefix}{URL: }
\providecommand{\Pubmedprefix}{pmid:}
\providecommand{\doi}[1]{\href{http://dx.doi.org/#1}{\path{#1}}}
\providecommand{\Pubmed}[1]{\href{pmid:#1}{\path{#1}}}
\providecommand{\bibinfo}[2]{#2}
\ifx\xfnm\relax \def\xfnm[#1]{\unskip,\space#1}\fi
\bibitem[{Cai et~al.(2023)Cai, Song, Cho, Wang, Wang, Yu, Liu, and Yu}]{userengagingheadline}
\bibinfo{author}{P.~Cai}, \bibinfo{author}{K.~Song}, \bibinfo{author}{S.~Cho}, \bibinfo{author}{H.~Wang}, \bibinfo{author}{X.~Wang}, \bibinfo{author}{H.~Yu}, \bibinfo{author}{F.~Liu}, \bibinfo{author}{D.~Yu},
\newblock \bibinfo{title}{Generating user-engaging news headlines},
\newblock in: \bibinfo{booktitle}{Proceedings of the 61st Annual Meeting of the Association for Computational Linguistics}, \bibinfo{year}{2023}, pp. \bibinfo{pages}{3265--3280}. \DOIprefix\doi{10.18653/v1/2023.acl-long.183}.
\bibitem[{De~Francisci~Morales et~al.(2012)De~Francisci~Morales, Gionis, and Lucchese}]{de2012chatter}
\bibinfo{author}{G.~De~Francisci~Morales}, \bibinfo{author}{A.~Gionis}, \bibinfo{author}{C.~Lucchese},
\newblock \bibinfo{title}{From chatter to headlines: harnessing the real-time web for personalized news recommendation},
\newblock in: \bibinfo{booktitle}{Proceedings of the fifth ACM international conference on Web search and data mining}, \bibinfo{year}{2012}, pp. \bibinfo{pages}{153--162}. \DOIprefix\doi{10.1145/2124295.2124315}.
\bibitem[{Koh et~al.(2022)Koh, Ju, Liu, and Pan}]{koh2022empirical}
\bibinfo{author}{H.~Y. Koh}, \bibinfo{author}{J.~Ju}, \bibinfo{author}{M.~Liu}, \bibinfo{author}{S.~Pan},
\newblock \bibinfo{title}{An empirical survey on long document summarization: Datasets, models, and metrics},
\newblock \bibinfo{journal}{ACM computing surveys} \bibinfo{volume}{55} (\bibinfo{year}{2022}) \bibinfo{pages}{1--35}. \DOIprefix\doi{10.1145/3545176}.
\bibitem[{Rao et~al.(2024)Rao, Aithal, and Singh}]{10.1145/3700639}
\bibinfo{author}{A.~Rao}, \bibinfo{author}{S.~Aithal}, \bibinfo{author}{S.~Singh},
\newblock \bibinfo{title}{Single-document abstractive text summarization: A systematic literature review},
\newblock \bibinfo{journal}{ACM Comput. Surv.} \bibinfo{volume}{57} (\bibinfo{year}{2024}). \DOIprefix\doi{10.1145/3700639}.
\bibitem[{Banerjee et~al.(2023)Banerjee, Mukherjee, Bandyopadhyay, and Pakray}]{BANERJEE2023103291}
\bibinfo{author}{S.~Banerjee}, \bibinfo{author}{S.~Mukherjee}, \bibinfo{author}{S.~Bandyopadhyay}, \bibinfo{author}{P.~Pakray},
\newblock \bibinfo{title}{An extract-then-abstract based method to generate disaster-news headlines using a dnn extractor followed by a transformer abstractor},
\newblock \bibinfo{journal}{Information Processing \& Management} \bibinfo{volume}{60} (\bibinfo{year}{2023}) \bibinfo{pages}{103291}. \DOIprefix\doi{10.1016/j.ipm.2023.103291}.
\bibitem[{Giarelis et~al.(2023)Giarelis, Mastrokostas, and Karacapilidis}]{app13137620}
\bibinfo{author}{N.~Giarelis}, \bibinfo{author}{C.~Mastrokostas}, \bibinfo{author}{N.~Karacapilidis},
\newblock \bibinfo{title}{Abstractive vs. extractive summarization: An experimental review},
\newblock \bibinfo{journal}{Applied Sciences} \bibinfo{volume}{13} (\bibinfo{year}{2023}). \DOIprefix\doi{10.3390/app13137620}.
\bibitem[{Alomari et~al.(2022)Alomari, Idris, Sabri, and Alsmadi}]{alomari2022deep}
\bibinfo{author}{A.~Alomari}, \bibinfo{author}{N.~Idris}, \bibinfo{author}{A.~Q.~M. Sabri}, \bibinfo{author}{I.~Alsmadi},
\newblock \bibinfo{title}{Deep reinforcement and transfer learning for abstractive text summarization: A review},
\newblock \bibinfo{journal}{Computer Speech \& Language} \bibinfo{volume}{71} (\bibinfo{year}{2022}) \bibinfo{pages}{101276}. \DOIprefix\doi{10.1016/j.csl.2021.101276}.
\bibitem[{El-Kassas et~al.(2021)El-Kassas, Salama, Rafea, and Mohamed}]{el2021automatic}
\bibinfo{author}{W.~S. El-Kassas}, \bibinfo{author}{C.~R. Salama}, \bibinfo{author}{A.~A. Rafea}, \bibinfo{author}{H.~K. Mohamed},
\newblock \bibinfo{title}{Automatic text summarization: A comprehensive survey},
\newblock \bibinfo{journal}{Expert systems with applications} \bibinfo{volume}{165} (\bibinfo{year}{2021}) \bibinfo{pages}{113679}. \DOIprefix\doi{10.1016/j.eswa.2020.113679}.
\bibitem[{Ahuir et~al.(2024)Ahuir, Gonzalez, Hurtado, and Segarra}]{app14020713}
\bibinfo{author}{V.~Ahuir}, \bibinfo{author}{J.-A. Gonzalez}, \bibinfo{author}{L.-F. Hurtado}, \bibinfo{author}{E.~Segarra},
\newblock \bibinfo{title}{Abstractive summarizers become emotional on news summarization},
\newblock \bibinfo{journal}{Applied Sciences} \bibinfo{volume}{14} (\bibinfo{year}{2024}). \DOIprefix\doi{10.3390/app14020713}.
\bibitem[{Ayana et~al.(2017)Ayana, Shen, Lin, Tu, Zhao, Liu, and Sun}]{Ayana2017}
\bibinfo{author}{Ayana}, \bibinfo{author}{S.-Q. Shen}, \bibinfo{author}{Y.-K. Lin}, \bibinfo{author}{C.-C. Tu}, \bibinfo{author}{Y.~Zhao}, \bibinfo{author}{Z.-Y. Liu}, \bibinfo{author}{M.-S. Sun},
\newblock \bibinfo{title}{Recent advances on neural headline generation},
\newblock \bibinfo{journal}{Journal of Computer Science and Technology} \bibinfo{volume}{32} (\bibinfo{year}{2017}) \bibinfo{pages}{768--784}. \DOIprefix\doi{10.1007/s11390-017-1758-3}.
\bibitem[{Hagar and Diakopoulos(2019)}]{hagar2019optimizing}
\bibinfo{author}{N.~Hagar}, \bibinfo{author}{N.~Diakopoulos},
\newblock \bibinfo{title}{Optimizing content with a/b headline testing: Changing newsroom practices},
\newblock \bibinfo{journal}{Media and Communication} \bibinfo{volume}{7} (\bibinfo{year}{2019}) \bibinfo{pages}{117--127}.
\bibitem[{Banerjee and Urminsky(2024)}]{banerjee2024language}
\bibinfo{author}{A.~Banerjee}, \bibinfo{author}{O.~Urminsky},
\newblock \bibinfo{title}{The language that drives engagement: A systematic large-scale analysis of headline experiments},
\newblock \bibinfo{journal}{Marketing Science}  (\bibinfo{year}{2024}).
\bibitem[{Akash et~al.(2023)Akash, Nayeem, Shohan, and Islam}]{akash-etal-2023-shironaam}
\bibinfo{author}{A.~U. Akash}, \bibinfo{author}{M.~T. Nayeem}, \bibinfo{author}{F.~T. Shohan}, \bibinfo{author}{T.~Islam},
\newblock \bibinfo{title}{Shironaam: {B}engali news headline generation using auxiliary information},
\newblock in: \bibinfo{booktitle}{Proceedings of the 17th Conference of the European Chapter of the Association for Computational Linguistics}, \bibinfo{publisher}{Association for Computational Linguistics}, \bibinfo{address}{Dubrovnik, Croatia}, \bibinfo{year}{2023}, pp. \bibinfo{pages}{52--67}. \URLprefix \url{https://aclanthology.org/2023.eacl-main.4}.
\bibitem[{Saad et~al.(2024)Saad, Mahi, Salim, and Hossain}]{SAAD2024110874}
\bibinfo{author}{A.~M. Saad}, \bibinfo{author}{U.~N. Mahi}, \bibinfo{author}{M.~S. Salim}, \bibinfo{author}{S.~I. Hossain},
\newblock \bibinfo{title}{Bangla news article dataset},
\newblock \bibinfo{journal}{Data in Brief} \bibinfo{volume}{57} (\bibinfo{year}{2024}) \bibinfo{pages}{110874}. \URLprefix \url{10.1016/j.dib.2024.110874}.
\bibitem[{Eberhard et~al.(2024)Eberhard, Simons, and Fennig}]{statb}
\bibinfo{author}{D.~M. Eberhard}, \bibinfo{author}{G.~F. Simons}, \bibinfo{author}{C.~D. Fennig}, \bibinfo{title}{Ethnologue: Languages of the world. twenty-seventh edition.}, \bibinfo{howpublished}{\url{https://www.ethnologue.com/}}, \bibinfo{year}{2024}. \bibinfo{note}{[last accessed 30 December 2024]}.
\bibitem[{Shaibani and Elnagar(2024)}]{servey}
\bibinfo{author}{A.~Y. Shaibani}, \bibinfo{author}{A.~M. Elnagar},
\newblock \bibinfo{title}{A survey of text summarization and headline generation methods in arabic: A survey of text summarization and headline generation methods in arabic},
\newblock in: \bibinfo{booktitle}{Proceedings of the 2024 9th International Conference on Machine Learning Technologies}, ICMLT '24, \bibinfo{publisher}{Association for Computing Machinery}, \bibinfo{address}{New York, NY, USA}, \bibinfo{year}{2024}, p. \bibinfo{pages}{317–323}. \DOIprefix\doi{10.1145/3674029.3674078}.
\bibitem[{Zeyad and Biradar(2024)}]{zeyad2024advancements}
\bibinfo{author}{A.~M.~A. Zeyad}, \bibinfo{author}{A.~Biradar},
\newblock \bibinfo{title}{Advancements in the efficacy of flan-t5 for abstractive text summarization: A multi-dataset evaluation using rouge and bertscore},
\newblock \bibinfo{journal}{2024 International Conference on Advancements in Power, Communication and Intelligent Systems (APCI)}  (\bibinfo{year}{2024}) \bibinfo{pages}{1--5}. \URLprefix \url{https://api.semanticscholar.org/CorpusID:271748003}.
\bibitem[{Yadav et~al.(2023)Yadav, Ranvijay, Yadav, and Maurya}]{Yadav2023}
\bibinfo{author}{A.~K. Yadav}, \bibinfo{author}{Ranvijay}, \bibinfo{author}{R.~S. Yadav}, \bibinfo{author}{A.~K. Maurya},
\newblock \bibinfo{title}{State-of-the-art approach to extractive text summarization: a comprehensive review},
\newblock \bibinfo{journal}{Multimedia Tools and Applications} \bibinfo{volume}{82} (\bibinfo{year}{2023}) \bibinfo{pages}{29135--29197}. \DOIprefix\doi{10.1007/s11042-023-14613-9}.
\bibitem[{Bharathi~Mohan et~al.(2023)Bharathi~Mohan, Prasanna~Kumar, Parathasarathy, Aravind, Hanish, and Pavithria}]{BharathiMohan2023}
\bibinfo{author}{G.~Bharathi~Mohan}, \bibinfo{author}{R.~Prasanna~Kumar}, \bibinfo{author}{S.~Parathasarathy}, \bibinfo{author}{S.~Aravind}, \bibinfo{author}{K.~B. Hanish}, \bibinfo{author}{G.~Pavithria}, \bibinfo{title}{Text Summarization for Big Data Analytics: A Comprehensive Review of GPT 2 and BERT Approaches}, \bibinfo{publisher}{Springer Nature Switzerland}, \bibinfo{address}{Cham}, \bibinfo{year}{2023}, pp. \bibinfo{pages}{247--264}. \DOIprefix\doi{10.1007/978-3-031-33808-3_14}.
\bibitem[{Cajueiro et~al.(2023)Cajueiro, Nery, Tavares, Melo, dos Reis, Weigang, and Celestino}]{cajueiro2023}
\bibinfo{author}{D.~O. Cajueiro}, \bibinfo{author}{A.~G. Nery}, \bibinfo{author}{I.~Tavares}, \bibinfo{author}{M.~K.~D. Melo}, \bibinfo{author}{S.~A. dos Reis}, \bibinfo{author}{L.~Weigang}, \bibinfo{author}{V.~R.~R. Celestino}, \bibinfo{title}{A comprehensive review of automatic text summarization techniques: method, data, evaluation and coding}, \bibinfo{year}{2023}. \href{http://arxiv.org/abs/2301.03403}{{\tt arXiv:2301.03403}}.
\bibitem[{Liu et~al.(2018)Liu, Li, Zhu, Zhang, and Zong}]{reviewheadline}
\bibinfo{author}{T.~Liu}, \bibinfo{author}{H.~Li}, \bibinfo{author}{J.~Zhu}, \bibinfo{author}{J.~Zhang}, \bibinfo{author}{C.~Zong},
\newblock \bibinfo{title}{Review headline generation with user embedding},
\newblock in: \bibinfo{editor}{M.~Sun}, \bibinfo{editor}{T.~Liu}, \bibinfo{editor}{X.~Wang}, \bibinfo{editor}{Z.~Liu}, \bibinfo{editor}{Y.~Liu} (Eds.), \bibinfo{booktitle}{Chinese Computational Linguistics and Natural Language Processing Based on Naturally Annotated Big Data}, \bibinfo{publisher}{Springer International Publishing}, \bibinfo{address}{Cham}, \bibinfo{year}{2018}, pp. \bibinfo{pages}{324--334}. \DOIprefix\doi{10.1007/978-3-030-01716-3_27}.
\bibitem[{Salehin et~al.(2019)Salehin, Rafat, Khan, and Abujar}]{salehin2019generating}
\bibinfo{author}{M.~Salehin}, \bibinfo{author}{A.~Rafat}, \bibinfo{author}{F.~Khan}, \bibinfo{author}{S.~Abujar},
\newblock \bibinfo{title}{Generating bengali news headlines: An attentive approach with sequence-to-sequence networks},
\newblock in: \bibinfo{booktitle}{Proceedings of the 8th International Conference System Modeling and Advancement in Research Trends (SMART)}, \bibinfo{year}{2019}, pp. \bibinfo{pages}{256--261}. \DOIprefix\doi{10.1109/SMART46866.2019.9117554}.
\bibitem[{Hayat et~al.(2023)Hayat, Das, and Hoque}]{hayat2023abstractive}
\bibinfo{author}{S.~M. A.~I. Hayat}, \bibinfo{author}{A.~Das}, \bibinfo{author}{M.~Hoque},
\newblock \bibinfo{title}{Abstractive bengali text summarization using transformer-based learning},
\newblock in: \bibinfo{booktitle}{6th International Conference on Electrical Information and Communication Technology (EICT)}, \bibinfo{year}{2023}, pp. \bibinfo{pages}{1--6}. \DOIprefix\doi{10.1109/EICT61409.2023.10427906}.
\bibitem[{Kabir et~al.(2024)Kabir, Islam, Laskar, Nayeem, Bari, and Hoque}]{kabir-etal-2024-benllm}
\bibinfo{author}{M.~Kabir}, \bibinfo{author}{M.~S. Islam}, \bibinfo{author}{M.~T.~R. Laskar}, \bibinfo{author}{M.~T. Nayeem}, \bibinfo{author}{M.~S. Bari}, \bibinfo{author}{E.~Hoque},
\newblock \bibinfo{title}{{B}en{LLM}-eval: A comprehensive evaluation into the potentials and pitfalls of large language models on {B}engali {NLP}},
\newblock in: \bibinfo{editor}{N.~Calzolari}, \bibinfo{editor}{M.-Y. Kan}, \bibinfo{editor}{V.~Hoste}, \bibinfo{editor}{A.~Lenci}, \bibinfo{editor}{S.~Sakti}, \bibinfo{editor}{N.~Xue} (Eds.), \bibinfo{booktitle}{Proceedings of the 2024 Joint International Conference on Computational Linguistics, Language Resources and Evaluation (LREC-COLING 2024)}, \bibinfo{publisher}{ELRA and ICCL}, \bibinfo{address}{Torino, Italia}, \bibinfo{year}{2024}, pp. \bibinfo{pages}{2238--2252}. \URLprefix \url{https://aclanthology.org/2024.lrec-main.201}.
\bibitem[{Grusky et~al.(2018)Grusky, Naaman, and Artzi}]{newsroom}
\bibinfo{author}{M.~Grusky}, \bibinfo{author}{M.~Naaman}, \bibinfo{author}{Y.~Artzi},
\newblock \bibinfo{title}{{N}ewsroom: A dataset of 1.3 million summaries with diverse extractive strategies},
\newblock in: \bibinfo{editor}{M.~Walker}, \bibinfo{editor}{H.~Ji}, \bibinfo{editor}{A.~Stent} (Eds.), \bibinfo{booktitle}{Proceedings of the 2018 Conference of the North {A}merican Chapter of the Association for Computational Linguistics: Human Language Technologies, Volume 1 (Long Papers)}, \bibinfo{publisher}{Association for Computational Linguistics}, \bibinfo{address}{New Orleans, Louisiana}, \bibinfo{year}{2018}, pp. \bibinfo{pages}{708--719}. \DOIprefix\doi{10.18653/v1/N18-1065}.
\bibitem[{Nallapati et~al.(2016)Nallapati, Zhou, dos santos, Gulcehre, and Xiang}]{sarvanidigital}
\bibinfo{author}{R.~Nallapati}, \bibinfo{author}{B.~Zhou}, \bibinfo{author}{C.~N. dos santos}, \bibinfo{author}{C.~Gulcehre}, \bibinfo{author}{B.~Xiang},
\newblock \bibinfo{title}{Abstractive text summarization using sequence-to-sequence rnns and beyond},
\newblock in: \bibinfo{booktitle}{Proceedings of the 20th SIGNLL Conference on Computational Natural Language Learning}, \bibinfo{year}{2016}, pp. \bibinfo{pages}{280--290}. \DOIprefix\doi{10.18653/v1/K16-1028}.
\bibitem[{Lins et~al.(2019)Lins, Oliveira, Cabral, Batista, Tenorio, Ferreira, Lima, de~Fran\c{c}a Pereira~e Silva, and Simske}]{cnncorpus}
\bibinfo{author}{R.~D. Lins}, \bibinfo{author}{H.~Oliveira}, \bibinfo{author}{L.~Cabral}, \bibinfo{author}{J.~Batista}, \bibinfo{author}{B.~Tenorio}, \bibinfo{author}{R.~Ferreira}, \bibinfo{author}{R.~Lima}, \bibinfo{author}{G.~de~Fran\c{c}a Pereira~e Silva}, \bibinfo{author}{S.~J. Simske},
\newblock \bibinfo{title}{The cnn-corpus: A large textual corpus for single-document extractive summarization},
\newblock in: \bibinfo{booktitle}{Proceedings of the ACM Symposium on Document Engineering 2019}, DocEng '19, \bibinfo{publisher}{Association for Computing Machinery}, \bibinfo{address}{New York, NY, USA}, \bibinfo{year}{2019}, pp. \bibinfo{pages}{1--10}. \DOIprefix\doi{10.1145/3342558.3345388}.
\bibitem[{Jiang and Dreyer(2024)}]{jiang-dreyer-2024-ccsum}
\bibinfo{author}{X.~Jiang}, \bibinfo{author}{M.~Dreyer},
\newblock \bibinfo{title}{{CCS}um: A large-scale and high-quality dataset for abstractive news summarization},
\newblock in: \bibinfo{editor}{K.~Duh}, \bibinfo{editor}{H.~Gomez}, \bibinfo{editor}{S.~Bethard} (Eds.), \bibinfo{booktitle}{Proceedings of the 2024 Conference of the North American Chapter of the Association for Computational Linguistics: Human Language Technologies (Volume 1: Long Papers)}, \bibinfo{publisher}{Association for Computational Linguistics}, \bibinfo{address}{Mexico City, Mexico}, \bibinfo{year}{2024}, pp. \bibinfo{pages}{7306--7336}. \DOIprefix\doi{10.18653/v1/2024.naacl-long.406}.
\bibitem[{Narayan et~al.(2018)Narayan, Cohen, and Lapata}]{xsum-dataset}
\bibinfo{author}{S.~Narayan}, \bibinfo{author}{S.~B. Cohen}, \bibinfo{author}{M.~Lapata},
\newblock \bibinfo{title}{Don{'}t give me the details, just the summary! topic-aware convolutional neural networks for extreme summarization},
\newblock in: \bibinfo{editor}{E.~Riloff}, \bibinfo{editor}{D.~Chiang}, \bibinfo{editor}{J.~Hockenmaier}, \bibinfo{editor}{J.~Tsujii} (Eds.), \bibinfo{booktitle}{Proceedings of the 2018 Conference on Empirical Methods in Natural Language Processing}, \bibinfo{publisher}{Association for Computational Linguistics}, \bibinfo{address}{Brussels, Belgium}, \bibinfo{year}{2018}, pp. \bibinfo{pages}{1797--1807}. \DOIprefix\doi{10.18653/v1/D18-1206}.
\bibitem[{Gu et~al.(2020)Gu, Mao, Han, Liu, Wu, Yu, Finnie, Yu, Zhai, and Zukoski}]{representativeheadlines}
\bibinfo{author}{X.~Gu}, \bibinfo{author}{Y.~Mao}, \bibinfo{author}{J.~Han}, \bibinfo{author}{J.~Liu}, \bibinfo{author}{Y.~Wu}, \bibinfo{author}{C.~Yu}, \bibinfo{author}{D.~Finnie}, \bibinfo{author}{H.~Yu}, \bibinfo{author}{J.~Zhai}, \bibinfo{author}{N.~Zukoski},
\newblock \bibinfo{title}{Generating representative headlines for news stories},
\newblock in: \bibinfo{booktitle}{Proceedings of The Web Conference 2020}, WWW '20, \bibinfo{publisher}{Association for Computing Machinery}, \bibinfo{address}{New York, NY, USA}, \bibinfo{year}{2020}, p. \bibinfo{pages}{1773–1784}. \DOIprefix\doi{10.1145/3366423.3380247}.
\bibitem[{Ao et~al.(2021)Ao, Wang, Luo, Qiao, He, and Xie}]{pens}
\bibinfo{author}{X.~Ao}, \bibinfo{author}{X.~Wang}, \bibinfo{author}{L.~Luo}, \bibinfo{author}{Y.~Qiao}, \bibinfo{author}{Q.~He}, \bibinfo{author}{X.~Xie},
\newblock \bibinfo{title}{{PENS}: A dataset and generic framework for personalized news headline generation},
\newblock in: \bibinfo{editor}{C.~Zong}, \bibinfo{editor}{F.~Xia}, \bibinfo{editor}{W.~Li}, \bibinfo{editor}{R.~Navigli} (Eds.), \bibinfo{booktitle}{Proceedings of the 59th Annual Meeting of the Association for Computational Linguistics and the 11th International Joint Conference on Natural Language Processing (Volume 1: Long Papers)}, \bibinfo{publisher}{Association for Computational Linguistics}, \bibinfo{address}{Online}, \bibinfo{year}{2021}, pp. \bibinfo{pages}{82--92}. \DOIprefix\doi{10.18653/v1/2021.acl-long.7}.
\bibitem[{Ao et~al.(2023)Ao, Luo, Wang, Yang, Chen, Qiao, He, and Xie}]{putyourvoice}
\bibinfo{author}{X.~Ao}, \bibinfo{author}{L.~Luo}, \bibinfo{author}{X.~Wang}, \bibinfo{author}{Z.~Yang}, \bibinfo{author}{J.-H. Chen}, \bibinfo{author}{Y.~Qiao}, \bibinfo{author}{Q.~He}, \bibinfo{author}{X.~Xie},
\newblock \bibinfo{title}{Put your voice on stage: Personalized headline generation for news articles},
\newblock \bibinfo{journal}{ACM Transactions on Knowledge Discovery from Data} \bibinfo{volume}{18} (\bibinfo{year}{2023}). \DOIprefix\doi{10.1145/3629168}.
\bibitem[{Jin et~al.(2020)Jin, Jin, Zhou, Orii, and Szolovits}]{jin2020hooksheadlinelearninggenerate}
\bibinfo{author}{D.~Jin}, \bibinfo{author}{Z.~Jin}, \bibinfo{author}{J.~T. Zhou}, \bibinfo{author}{L.~Orii}, \bibinfo{author}{P.~Szolovits},
\newblock \bibinfo{title}{Hooks in the headline: Learning to generate headlines with controlled styles},
\newblock in: \bibinfo{editor}{D.~Jurafsky}, \bibinfo{editor}{J.~Chai}, \bibinfo{editor}{N.~Schluter}, \bibinfo{editor}{J.~Tetreault} (Eds.), \bibinfo{booktitle}{Proceedings of the 58th Annual Meeting of the Association for Computational Linguistics}, \bibinfo{publisher}{Association for Computational Linguistics}, \bibinfo{address}{Online}, \bibinfo{year}{2020}, pp. \bibinfo{pages}{5082--5093}. \DOIprefix\doi{10.18653/v1/2020.acl-main.456}.
\bibitem[{Sandhaus(2008)}]{AB2/GZC6PL_2008}
\bibinfo{author}{E.~Sandhaus}, \bibinfo{title}{{The New York Times Annotated Corpus}}, \bibinfo{year}{2008}. \URLprefix \url{https://hdl.handle.net/11272.1/AB2/GZC6PL}. \DOIprefix\doi{11272.1/AB2/GZC6PL}.
\bibitem[{Takase et~al.(2016)Takase, Suzuki, Okazaki, Hirao, and Nagata}]{takase2016neural}
\bibinfo{author}{S.~Takase}, \bibinfo{author}{J.~Suzuki}, \bibinfo{author}{N.~Okazaki}, \bibinfo{author}{T.~Hirao}, \bibinfo{author}{M.~Nagata},
\newblock \bibinfo{title}{Neural headline generation on {A}bstract {M}eaning {R}epresentation},
\newblock in: \bibinfo{editor}{J.~Su}, \bibinfo{editor}{K.~Duh}, \bibinfo{editor}{X.~Carreras} (Eds.), \bibinfo{booktitle}{Proceedings of the 2016 Conference on Empirical Methods in Natural Language Processing}, \bibinfo{publisher}{Association for Computational Linguistics}, \bibinfo{address}{Austin, Texas}, \bibinfo{year}{2016}, pp. \bibinfo{pages}{1054--1059}. \DOIprefix\doi{10.18653/v1/D16-1112}.
\bibitem[{{National Institute of Standard and Technology}(2014)}]{duc}
\bibinfo{author}{{National Institute of Standard and Technology}}, \bibinfo{title}{Document understanding conferences}, \bibinfo{howpublished}{\url{https://www-nlpir.nist.gov/projects/duc/data.html}}, \bibinfo{year}{2014}. \bibinfo{note}{Accessed: 30 December 2024}.
\bibitem[{Graff and Cieri(2003)}]{english-gigaword}
\bibinfo{author}{D.~Graff}, \bibinfo{author}{C.~Cieri}, \bibinfo{title}{English gigaword}, \bibinfo{howpublished}{\url{https://catalog.ldc.upenn.edu/LDC2003T05}}, \bibinfo{year}{2003}. \DOIprefix\doi{10.35111/0z6y-q265}.
\bibitem[{Napoles et~al.(2012)Napoles, Gormley, and Van~Durme}]{gigaword}
\bibinfo{author}{C.~Napoles}, \bibinfo{author}{M.~Gormley}, \bibinfo{author}{B.~Van~Durme},
\newblock \bibinfo{title}{Annotated gigaword},
\newblock in: \bibinfo{booktitle}{Proceedings of the Joint Workshop on Automatic Knowledge Base Construction and Web-Scale Knowledge Extraction}, AKBC-WEKEX '12, \bibinfo{publisher}{Association for Computational Linguistics}, \bibinfo{year}{2012}, p. \bibinfo{pages}{95–100}. \URLprefix \url{https://aclanthology.org/W12-3018.pdf}.
\bibitem[{Singh et~al.(2021)Singh, Khetarpaul, Gorantla, and Allada}]{singh2021sheg}
\bibinfo{author}{R.~K. Singh}, \bibinfo{author}{S.~Khetarpaul}, \bibinfo{author}{R.~Gorantla}, \bibinfo{author}{S.~G. Allada},
\newblock \bibinfo{title}{Sheg: summarization and headline generation of news articles using deep learning},
\newblock \bibinfo{journal}{Neural Computing and Applications} \bibinfo{volume}{33} (\bibinfo{year}{2021}) \bibinfo{pages}{3251--3265}. \DOIprefix\doi{10.1007/s00521-020-05188-9}.
\bibitem[{Sen and Yanikoglu(2018)}]{suder}
\bibinfo{author}{M.~Sen}, \bibinfo{author}{B.~Yanikoglu},
\newblock \bibinfo{title}{Document classification of suder turkish news corpora},
\newblock in: \bibinfo{booktitle}{Proceedings of the 2018 26th Signal Processing and Communications Applications Conference (SIU)}, \bibinfo{year}{2018}, pp. \bibinfo{pages}{1--4}. \DOIprefix\doi{10.1109/SIU.2018.8404790}.
\bibitem[{Ogunremi et~al.(2024)Ogunremi, sessi Akojenu, Soronnadi, Adekanmbi, and Adelani}]{ogunremi2024afrihg}
\bibinfo{author}{T.~Ogunremi}, \bibinfo{author}{S.~sessi Akojenu}, \bibinfo{author}{A.~Soronnadi}, \bibinfo{author}{O.~Adekanmbi}, \bibinfo{author}{D.~I. Adelani},
\newblock \bibinfo{title}{Afri{HG}: News headline generation for african languages},
\newblock in: \bibinfo{booktitle}{5th Workshop on African Natural Language Processing}, \bibinfo{year}{2024}, p.~\bibinfo{pages}{4}. \URLprefix \url{https://openreview.net/forum?id=fw7g7pNUDl}.
\bibitem[{Bukhtiyarov and Gusev(2020)}]{bukhtiyarov}
\bibinfo{author}{A.~Bukhtiyarov}, \bibinfo{author}{I.~Gusev}, \bibinfo{title}{Advances of Transformer-Based Models for News Headline Generation}, \bibinfo{publisher}{Springer}, \bibinfo{year}{2020}, pp. \bibinfo{pages}{54--61}. \DOIprefix\doi{10.1007/978-3-030-59082-6_4}.
\bibitem[{Gavrilov et~al.(2019)Gavrilov, Kalaidin, and Malykh}]{gavrilov2019self}
\bibinfo{author}{D.~Gavrilov}, \bibinfo{author}{P.~Kalaidin}, \bibinfo{author}{V.~Malykh}, \bibinfo{title}{Self-attentive model for headline generation}, \bibinfo{year}{2019}. \DOIprefix\doi{10.1007/978-3-030-15719-7_11}.
\bibitem[{Yutkin(2019)}]{lenta}
\bibinfo{author}{D.~Yutkin}, \bibinfo{title}{Lenta}, \bibinfo{howpublished}{\url{https://github.com/yutkin/Lenta.Ru-News-Dataset}}, \bibinfo{year}{2019}. \bibinfo{note}{Accessed: 30 December 2024}.
\bibitem[{Hu et~al.(2015)Hu, Chen, and Zhu}]{hu-etal-2015-lcsts}
\bibinfo{author}{B.~Hu}, \bibinfo{author}{Q.~Chen}, \bibinfo{author}{F.~Zhu},
\newblock \bibinfo{title}{{LCSTS}: A large scale {C}hinese short text summarization dataset},
\newblock in: \bibinfo{editor}{L.~M{\`a}rquez}, \bibinfo{editor}{C.~Callison-Burch}, \bibinfo{editor}{J.~Su} (Eds.), \bibinfo{booktitle}{Proceedings of the 2015 Conference on Empirical Methods in Natural Language Processing}, \bibinfo{publisher}{Association for Computational Linguistics}, \bibinfo{address}{Lisbon, Portugal}, \bibinfo{year}{2015}, pp. \bibinfo{pages}{1967--1972}. \DOIprefix\doi{10.18653/v1/D15-1229}.
\bibitem[{Madasu et~al.(2023)Madasu, Kanumolu, Surange, and Shrivastava}]{madasu-etal-2023-mukhyansh}
\bibinfo{author}{L.~Madasu}, \bibinfo{author}{G.~Kanumolu}, \bibinfo{author}{N.~Surange}, \bibinfo{author}{M.~Shrivastava},
\newblock \bibinfo{title}{Mukhyansh: A headline generation dataset for {I}ndic languages},
\newblock in: \bibinfo{editor}{C.-R. Huang}, \bibinfo{editor}{Y.~Harada}, \bibinfo{editor}{J.-B. Kim}, \bibinfo{editor}{S.~Chen}, \bibinfo{editor}{Y.-Y. Hsu}, \bibinfo{editor}{E.~Chersoni}, \bibinfo{editor}{P.~A}, \bibinfo{editor}{W.~H. Zeng}, \bibinfo{editor}{B.~Peng}, \bibinfo{editor}{Y.~Li}, \bibinfo{editor}{J.~Li} (Eds.), \bibinfo{booktitle}{Proceedings of the 37th Pacific Asia Conference on Language, Information and Computation}, \bibinfo{publisher}{Association for Computational Linguistics}, \bibinfo{address}{Hong Kong, China}, \bibinfo{year}{2023}, pp. \bibinfo{pages}{620--634}. \URLprefix \url{https://aclanthology.org/2023.paclic-1.62}.
\bibitem[{Aralikatte et~al.(2023)Aralikatte, Cheng, Doddapaneni, and Cheung}]{aralikatte-etal-2023-varta}
\bibinfo{author}{R.~Aralikatte}, \bibinfo{author}{Z.~Cheng}, \bibinfo{author}{S.~Doddapaneni}, \bibinfo{author}{J.~C.~K. Cheung},
\newblock \bibinfo{title}{Varta: A large-scale headline-generation dataset for {I}ndic languages},
\newblock in: \bibinfo{editor}{A.~Rogers}, \bibinfo{editor}{J.~Boyd-Graber}, \bibinfo{editor}{N.~Okazaki} (Eds.), \bibinfo{booktitle}{Findings of the Association for Computational Linguistics: ACL 2023}, \bibinfo{publisher}{Association for Computational Linguistics}, \bibinfo{address}{Toronto, Canada}, \bibinfo{year}{2023}, pp. \bibinfo{pages}{3468--3492}. \DOIprefix\doi{10.18653/v1/2023.findings-acl.215}.
\bibitem[{Li et~al.(2021)Li, Yu, Chen, and Guo}]{9507422}
\bibinfo{author}{P.~Li}, \bibinfo{author}{J.~Yu}, \bibinfo{author}{J.~Chen}, \bibinfo{author}{B.~Guo},
\newblock \bibinfo{title}{Hg-news: News headline generation based on a generative pre-training model},
\newblock \bibinfo{journal}{IEEE Access} \bibinfo{volume}{9} (\bibinfo{year}{2021}) \bibinfo{pages}{110039--110046}. \DOIprefix\doi{10.1109/ACCESS.2021.3102741}.
\bibitem[{Hasan et~al.(2021)Hasan, Bhattacharjee, Islam, Mubasshir, Li, Kang, Rahman, and Shahriyar}]{hasan2021xlsumlargescalemultilingualabstractive}
\bibinfo{author}{T.~Hasan}, \bibinfo{author}{A.~Bhattacharjee}, \bibinfo{author}{M.~S. Islam}, \bibinfo{author}{K.~Mubasshir}, \bibinfo{author}{Y.-F. Li}, \bibinfo{author}{Y.-B. Kang}, \bibinfo{author}{M.~S. Rahman}, \bibinfo{author}{R.~Shahriyar},
\newblock \bibinfo{title}{{XL-Sum}: Large-scale multilingual abstractive summarization for 44 languages},
\newblock in: \bibinfo{booktitle}{The Joint Conference of the 59th Annual Meeting of the Association for Computational Linguistics and the 11th International Joint Conference on Natural Language Processing (ACL-IJCNLP)}, \bibinfo{publisher}{Association for Computational Linguistics}, \bibinfo{year}{2021}, pp. \bibinfo{pages}{4693--4703}.
\bibitem[{Ahmad et~al.(2022)Ahmad, Alqurashi, and Mehmood}]{potrika}
\bibinfo{author}{I.~Ahmad}, \bibinfo{author}{F.~Alqurashi}, \bibinfo{author}{R.~Mehmood}, \bibinfo{title}{Potrika: Raw and balanced newspaper datasets in the bangla language with eight topics and five attributes}, \bibinfo{year}{2022}. \href{http://arxiv.org/abs/2210.09389}{{\tt arXiv:2210.09389}}.
\bibitem[{Karaca and Aydın(2023)}]{turkisnews}
\bibinfo{author}{A.~Karaca}, \bibinfo{author}{O.~Aydın},
\newblock \bibinfo{title}{Generating headlines for turkish news texts with transformer architecture based deep learning method},
\newblock \bibinfo{journal}{Gazi Üniversitesi Mühendislik-Mimarlık Fakültesi Dergisi} \bibinfo{volume}{39} (\bibinfo{year}{2023}) \bibinfo{pages}{485--495}. \DOIprefix\doi{10.17341/gazimmfd.963240}.
\bibitem[{Theledi and Pule(2024)}]{theledi2024president}
\bibinfo{author}{K.~Theledi}, \bibinfo{author}{V.~M. Pule},
\newblock \bibinfo{title}{President's speech and terminology used during the covid-19 pandemic: The interpretation of linguistic meaning in context and situational context},
\newblock in: \bibinfo{booktitle}{Public Health Communication Challenges to Minority and Indigenous Communities}, \bibinfo{publisher}{IGI Global}, \bibinfo{year}{2024}, pp. \bibinfo{pages}{92--107}.
\bibitem[{Liu et~al.(2020)Liu, Gong, Yan, Fu, Shao, Jiang, Lv, and Duan}]{liu-etal-2020-diverse}
\bibinfo{author}{D.~Liu}, \bibinfo{author}{Y.~Gong}, \bibinfo{author}{Y.~Yan}, \bibinfo{author}{J.~Fu}, \bibinfo{author}{B.~Shao}, \bibinfo{author}{D.~Jiang}, \bibinfo{author}{J.~Lv}, \bibinfo{author}{N.~Duan},
\newblock \bibinfo{title}{Diverse, controllable, and keyphrase-aware: A corpus and method for news multi-headline generation},
\newblock in: \bibinfo{editor}{B.~Webber}, \bibinfo{editor}{T.~Cohn}, \bibinfo{editor}{Y.~He}, \bibinfo{editor}{Y.~Liu} (Eds.), \bibinfo{booktitle}{Proceedings of the 2020 Conference on Empirical Methods in Natural Language Processing (EMNLP)}, \bibinfo{publisher}{Association for Computational Linguistics}, \bibinfo{address}{Online}, \bibinfo{year}{2020}, pp. \bibinfo{pages}{6241--6250}. \DOIprefix\doi{10.18653/v1/2020.emnlp-main.505}.
\bibitem[{Kiefer(2022)}]{kiefer2022case}
\bibinfo{author}{S.~Kiefer},
\newblock \bibinfo{title}{Case: Explaining text classifications by fusion of local surrogate explanation models with contextual and semantic knowledge},
\newblock \bibinfo{journal}{Information Fusion} \bibinfo{volume}{77} (\bibinfo{year}{2022}) \bibinfo{pages}{184--195}. \DOIprefix\doi{10.1016/j.inffus.2021.07.014}.
\bibitem[{Chen(2019)}]{chen2019ci}
\bibinfo{author}{N.~Chen},
\newblock \bibinfo{title}{Ci-snf: Exploiting contextual information to improve snf based information retrieval},
\newblock \bibinfo{journal}{Information Fusion} \bibinfo{volume}{52} (\bibinfo{year}{2019}) \bibinfo{pages}{175--186}. \DOIprefix\doi{j.inffus.2018.08.004}.
\bibitem[{Zhu et~al.(2023)Zhu, Zhu, Zhang, Xu, and Kong}]{zhu2023multimodal}
\bibinfo{author}{L.~Zhu}, \bibinfo{author}{Z.~Zhu}, \bibinfo{author}{C.~Zhang}, \bibinfo{author}{Y.~Xu}, \bibinfo{author}{X.~Kong},
\newblock \bibinfo{title}{Multimodal sentiment analysis based on fusion methods: A survey},
\newblock \bibinfo{journal}{Information Fusion} \bibinfo{volume}{95} (\bibinfo{year}{2023}) \bibinfo{pages}{306--325}. \DOIprefix\doi{j.inffus.2023.02.028}.
\bibitem[{Aziz et~al.(2023)Aziz, Chowdhury, Kabir, Chy, and Siddique}]{aziz2023mmtf}
\bibinfo{author}{A.~Aziz}, \bibinfo{author}{N.~K. Chowdhury}, \bibinfo{author}{M.~A. Kabir}, \bibinfo{author}{A.~N. Chy}, \bibinfo{author}{M.~J. Siddique}, \bibinfo{title}{Mmtf-des: A fusion of multimodal transformer models for desire, emotion, and sentiment analysis of social media data}, \bibinfo{year}{2023}. \href{http://arxiv.org/abs/2310.14143}{{\tt arXiv:2310.14143}}.
\bibitem[{Hasan et~al.(2020)Hasan, Bhattacharjee, Samin, Hasan, Basak, Rahman, and Shahriyar}]{hasan-etal-2020-low}
\bibinfo{author}{T.~Hasan}, \bibinfo{author}{A.~Bhattacharjee}, \bibinfo{author}{K.~Samin}, \bibinfo{author}{M.~Hasan}, \bibinfo{author}{M.~Basak}, \bibinfo{author}{M.~S. Rahman}, \bibinfo{author}{R.~Shahriyar},
\newblock \bibinfo{title}{Not low-resource anymore: Aligner ensembling, batch filtering, and new datasets for {B}engali-{E}nglish machine translation},
\newblock in: \bibinfo{booktitle}{Proceedings of the 2020 Conference on Empirical Methods in Natural Language Processing (EMNLP)}, \bibinfo{publisher}{Association for Computational Linguistics}, \bibinfo{year}{2020}, pp. \bibinfo{pages}{2612--2623}. \DOIprefix\doi{10.18653/v1/2020.emnlp-main.207}.
\bibitem[{Lin(2004)}]{lin2004rouge}
\bibinfo{author}{C.-Y. Lin},
\newblock \bibinfo{title}{{ROUGE}: A package for automatic evaluation of summaries},
\newblock in: \bibinfo{booktitle}{Text Summarization Branches Out}, \bibinfo{publisher}{Association for Computational Linguistics}, \bibinfo{address}{Barcelona, Spain}, \bibinfo{year}{2004}, pp. \bibinfo{pages}{74--81}. \URLprefix \url{https://aclanthology.org/W04-1013}.
\bibitem[{Papineni et~al.(2002)Papineni, Roukos, Ward, and Zhu}]{papineni2002bleu}
\bibinfo{author}{K.~Papineni}, \bibinfo{author}{S.~Roukos}, \bibinfo{author}{T.~Ward}, \bibinfo{author}{W.-J. Zhu},
\newblock \bibinfo{title}{{B}leu: a method for automatic evaluation of machine translation},
\newblock in: \bibinfo{editor}{P.~Isabelle}, \bibinfo{editor}{E.~Charniak}, \bibinfo{editor}{D.~Lin} (Eds.), \bibinfo{booktitle}{Proceedings of the 40th Annual Meeting of the Association for Computational Linguistics}, \bibinfo{publisher}{Association for Computational Linguistics}, \bibinfo{address}{Philadelphia, Pennsylvania, USA}, \bibinfo{year}{2002}, pp. \bibinfo{pages}{311--318}. \DOIprefix\doi{10.3115/1073083.1073135}.
\bibitem[{Banerjee and Lavie(2005)}]{banerjee2005meteor}
\bibinfo{author}{S.~Banerjee}, \bibinfo{author}{A.~Lavie},
\newblock \bibinfo{title}{{METEOR}: An automatic metric for {MT} evaluation with improved correlation with human judgments},
\newblock in: \bibinfo{editor}{J.~Goldstein}, \bibinfo{editor}{A.~Lavie}, \bibinfo{editor}{C.-Y. Lin}, \bibinfo{editor}{C.~Voss} (Eds.), \bibinfo{booktitle}{Proceedings of the {ACL} Workshop on Intrinsic and Extrinsic Evaluation Measures for Machine Translation and/or Summarization}, \bibinfo{publisher}{Association for Computational Linguistics}, \bibinfo{address}{Ann Arbor, Michigan}, \bibinfo{year}{2005}, pp. \bibinfo{pages}{65--72}. \URLprefix \url{https://aclanthology.org/W05-0909}.
\bibitem[{Zhang et~al.(2020)Zhang, Kishore, Wu, Weinberger, and Artzi}]{bertscore}
\bibinfo{author}{T.~Zhang}, \bibinfo{author}{V.~Kishore}, \bibinfo{author}{F.~Wu}, \bibinfo{author}{K.~Q. Weinberger}, \bibinfo{author}{Y.~Artzi},
\newblock \bibinfo{title}{{BERTScore}: Evaluating text generation with bert},
\newblock in: \bibinfo{booktitle}{International Conference on Learning Representations}, \bibinfo{year}{2020}, p.~\bibinfo{pages}{43}. \URLprefix \url{https://openreview.net/forum?id=SkeHuCVFDr}.
\bibitem[{Raffel et~al.(2020)Raffel, Shazeer, Roberts, Lee, Narang, Matena, Zhou, Li, and Liu}]{JMLR:v21:20-074}
\bibinfo{author}{C.~Raffel}, \bibinfo{author}{N.~Shazeer}, \bibinfo{author}{A.~Roberts}, \bibinfo{author}{K.~Lee}, \bibinfo{author}{S.~Narang}, \bibinfo{author}{M.~Matena}, \bibinfo{author}{Y.~Zhou}, \bibinfo{author}{W.~Li}, \bibinfo{author}{P.~J. Liu},
\newblock \bibinfo{title}{Exploring the limits of transfer learning with a unified text-to-text transformer},
\newblock \bibinfo{journal}{Journal of Machine Learning Research} \bibinfo{volume}{21} (\bibinfo{year}{2020}) \bibinfo{pages}{1--67}. \URLprefix \url{http://jmlr.org/papers/v21/20-074.html}.
\bibitem[{Lewis et~al.(2020)Lewis, Liu, Goyal, Ghazvininejad, Mohamed, Levy, Stoyanov, and Zettlemoyer}]{lewis2019bart}
\bibinfo{author}{M.~Lewis}, \bibinfo{author}{Y.~Liu}, \bibinfo{author}{N.~Goyal}, \bibinfo{author}{M.~Ghazvininejad}, \bibinfo{author}{A.~Mohamed}, \bibinfo{author}{O.~Levy}, \bibinfo{author}{V.~Stoyanov}, \bibinfo{author}{L.~Zettlemoyer},
\newblock \bibinfo{title}{{BART}: Denoising sequence-to-sequence pre-training for natural language generation, translation, and comprehension},
\newblock in: \bibinfo{editor}{D.~Jurafsky}, \bibinfo{editor}{J.~Chai}, \bibinfo{editor}{N.~Schluter}, \bibinfo{editor}{J.~Tetreault} (Eds.), \bibinfo{booktitle}{Proceedings of the 58th Annual Meeting of the Association for Computational Linguistics}, \bibinfo{publisher}{Association for Computational Linguistics}, \bibinfo{address}{Online}, \bibinfo{year}{2020}, pp. \bibinfo{pages}{7871--7880}. \DOIprefix\doi{10.18653/v1/2020.acl-main.703}.
\bibitem[{Bhattacharjee et~al.(2022)Bhattacharjee, Hasan, Ahmad, and Shahriyar}]{bhattacharjee2022banglanlg}
\bibinfo{author}{A.~Bhattacharjee}, \bibinfo{author}{T.~Hasan}, \bibinfo{author}{W.~U. Ahmad}, \bibinfo{author}{R.~Shahriyar}, \bibinfo{title}{{BanglaNLG}: Benchmarks and resources for evaluating low-resource natural language generation in bangla}, \bibinfo{year}{2022}. \href{http://arxiv.org/abs/2205.11081}{{\tt arXiv:2205.11081}}.
\bibitem[{Muennighoff et~al.(2023)Muennighoff, Wang, Sutawika, Roberts, Biderman, Le~Scao, Bari, Shen, Yong, Schoelkopf, Tang, Radev, Aji, Almubarak, Albanie, Alyafeai, Webson, Raff, and Raffel}]{muennighoff2022crosslingual}
\bibinfo{author}{N.~Muennighoff}, \bibinfo{author}{T.~Wang}, \bibinfo{author}{L.~Sutawika}, \bibinfo{author}{A.~Roberts}, \bibinfo{author}{S.~Biderman}, \bibinfo{author}{T.~Le~Scao}, \bibinfo{author}{M.~S. Bari}, \bibinfo{author}{S.~Shen}, \bibinfo{author}{Z.~X. Yong}, \bibinfo{author}{H.~Schoelkopf}, \bibinfo{author}{X.~Tang}, \bibinfo{author}{D.~Radev}, \bibinfo{author}{A.~F. Aji}, \bibinfo{author}{K.~Almubarak}, \bibinfo{author}{S.~Albanie}, \bibinfo{author}{Z.~Alyafeai}, \bibinfo{author}{A.~Webson}, \bibinfo{author}{E.~Raff}, \bibinfo{author}{C.~Raffel},
\newblock \bibinfo{title}{Crosslingual generalization through multitask finetuning},
\newblock in: \bibinfo{editor}{A.~Rogers}, \bibinfo{editor}{J.~Boyd-Graber}, \bibinfo{editor}{N.~Okazaki} (Eds.), \bibinfo{booktitle}{Proceedings of the 61st Annual Meeting of the Association for Computational Linguistics (Volume 1: Long Papers)}, \bibinfo{publisher}{Association for Computational Linguistics}, \bibinfo{address}{Toronto, Canada}, \bibinfo{year}{2023}, pp. \bibinfo{pages}{15991--16111}. \DOIprefix\doi{10.18653/v1/2023.acl-long.891}.
\bibitem[{Xue et~al.(2021)Xue, Constant, Roberts, Kale, Al-Rfou, Siddhant, Barua, and Raffel}]{xue2021mt5}
\bibinfo{author}{L.~Xue}, \bibinfo{author}{N.~Constant}, \bibinfo{author}{A.~Roberts}, \bibinfo{author}{M.~Kale}, \bibinfo{author}{R.~Al-Rfou}, \bibinfo{author}{A.~Siddhant}, \bibinfo{author}{A.~Barua}, \bibinfo{author}{C.~Raffel},
\newblock \bibinfo{title}{m{T}5: A massively multilingual pre-trained text-to-text transformer},
\newblock in: \bibinfo{editor}{K.~Toutanova}, \bibinfo{editor}{A.~Rumshisky}, \bibinfo{editor}{L.~Zettlemoyer}, \bibinfo{editor}{D.~Hakkani-Tur}, \bibinfo{editor}{I.~Beltagy}, \bibinfo{editor}{S.~Bethard}, \bibinfo{editor}{R.~Cotterell}, \bibinfo{editor}{T.~Chakraborty}, \bibinfo{editor}{Y.~Zhou} (Eds.), \bibinfo{booktitle}{Proceedings of the 2021 Conference of the North American Chapter of the Association for Computational Linguistics: Human Language Technologies}, \bibinfo{publisher}{Association for Computational Linguistics}, \bibinfo{address}{Online}, \bibinfo{year}{2021}, pp. \bibinfo{pages}{483--498}. \DOIprefix\doi{10.18653/v1/2021.naacl-main.41}.
\bibitem[{Tang et~al.(2021)Tang, Tran, Li, Chen, Goyal, Chaudhary, Gu, and Fan}]{tang2020multilingual}
\bibinfo{author}{Y.~Tang}, \bibinfo{author}{C.~Tran}, \bibinfo{author}{X.~Li}, \bibinfo{author}{P.-J. Chen}, \bibinfo{author}{N.~Goyal}, \bibinfo{author}{V.~Chaudhary}, \bibinfo{author}{J.~Gu}, \bibinfo{author}{A.~Fan},
\newblock \bibinfo{title}{Multilingual translation from denoising pre-training},
\newblock in: \bibinfo{booktitle}{The Joint Conference of the 59th Annual Meeting of the Association for Computational Linguistics and the 11th International Joint Conference on Natural Language Processing (ACL-IJCNLP)}, \bibinfo{publisher}{Association for Computational Linguistics}, \bibinfo{year}{2021}, pp. \bibinfo{pages}{3450--3466}.
\bibitem[{Rahman and Mamun(2024)}]{rahman2024rise}
\bibinfo{author}{A.~Rahman}, \bibinfo{author}{A.~Mamun},
\newblock \bibinfo{title}{The rise of clickbait headlines: A study on media platforms from bangladesh},
\newblock \bibinfo{journal}{Athens Journal of Mass Media and Communications} \bibinfo{volume}{10} (\bibinfo{year}{2024}) \bibinfo{pages}{109--130}. \DOIprefix\doi{10.30958/ajmmc.10-2-3}.

\end{thebibliography}
\end{document}